\documentclass{article}

\usepackage{authblk}
\usepackage{subfigure}
\usepackage{multirow} 
\usepackage{booktabs}
\usepackage{amsmath}
\usepackage{amsthm}
\usepackage{array}
\usepackage{amssymb}
\usepackage{cite}
\usepackage[numbers]{natbib}
\usepackage[ruled, vlined]{algorithm2e}
\usepackage{subfiles}
\usepackage[colorlinks, linkcolor=black, citecolor=black]{hyperref}
\usepackage{setspace}
\usepackage{enumitem}
\usepackage{graphicx}
\usepackage{xspace}
\usepackage{xcolor}
\usepackage{bm}
\usepackage{rotating}

\graphicspath{{figures/}{../figures/}}

\newcommand{\nop}[1]{}

\usepackage{xcolor}
\usepackage{color, soul}
\newcommand{\todo}[1]{\textcolor{red}{\hl{[#1]}}}

\newcolumntype{P}[1]{>{\centering\arraybackslash}p{#1}}

\newtheorem{example}{Example}

\begin{document}
\begin{sloppy}

\title{Model Complexity of Deep Learning: A Survey}
\author[1]{	Xia~Hu}
\author[2]{Lingyang~Chu}
\author[1]{	Jian~Pei}
\author[3]{	Weiqing~Liu}
\author[3]{Jiang~Bian}
\affil[1]{School of Computing Science, Simon Fraser University, Burnaby, Canada, \texttt{huxiah@sfu.ca, jpei@cs.sfu.ca}}
\affil[2]{Department of Computing and Software, McMaster University, Hamilton, Canada, \texttt{chul9@mcmaster.ca}}
\affil[3]{Microsoft Research, Beijing, China, \texttt{\{Weiqing.liu, Jiang.bian\}@microsoft.com}}



\maketitle

\begin{abstract}
	Model complexity is a fundamental problem in deep learning. In this paper we conduct a systematic overview of the latest studies on model complexity in deep learning. Model complexity of deep learning can be categorized into expressive capacity and effective model complexity.  We review the existing studies on those two categories along four important factors, including model framework, model size, optimization process and data complexity.  We also discuss the applications of deep learning model complexity including understanding model generalization, model optimization, and model selection and design.  We conclude by proposing several interesting future directions.
\end{abstract}


\section{Introduction}
\label{sec:intro}

Mainly due to its superior performance, deep learning is disruptive in many applications, such as computer vision~\cite{he2016deep}, natural language processing~\cite{lample2018phrase} and computational finance~\cite{rebentrost2018quantum}. At the same time, however, a series of fundamental questions about deep learning models remain, such as why deep learning can achieve substantially better expressive power comparing to classical machine learning models, 
how to understand and quantify the generalization capability of deep models, and how to understand and improve the optimization process. Model complexity of deep learning is a core problem and is related to many those fundamental questions.

Model complexity of deep learning is concerned about, for a certain deep learning architecture, how complicated  problems the deep learning model can express~\cite{bianchini2014complexity,hu2020lann,montufar2014number,raghu2017expressive}\nop{\todo{Replace hu2020lann by the KDD 2020 paper}}.  Understanding the complexity of a deep model is a key to precisely understanding the capability and limitation of the model. Exploring model complexity is necessary not only for understanding a deep model itself, but also for investigating many other related fundamental questions. 
For example, from the statistical learning theory point of view, the expressive power of a model is used to bound the generalization error~\cite{mohri2018foundations}. Some recent studies propose norm-based model complexity~\cite{liang2017fisher} and sensitivity-based model complexity~\cite{neyshabur2017exploring,novak2018sensitivity} to explore the generalization capability of deep models. Moreover, detecting changes of model complexity in a training process can provide insights into understanding and improving the performance of model optimization and regularization~\cite{hu2020lann,nakkiran2019deep,raghu2017expressive}.

The investigation of machine learning model complexity dates back to decades ago. A series of early studies in 1990s discuss the complexity of classical machine learning models~\cite{bohanec1994trading,buhrman2002complexity,bulso2019complexity,yao1997decision}. One representative model is decision tree~\cite{breiman1984classification}, whose complexity is always measured by tree depth~\cite{buhrman2002complexity} and number of leaf nodes~\cite{bohanec1994trading}. Another frequent subject in model complexity analysis is logistic regression, which is the foundation of a large number of parameterized models. The model complexity of logistic regression is investigated from the perspectives of Vapnik-Chervonenicks theory~\cite{cherkassky1999model,vapnik2013nature}, Rademacher complexity~\cite{kakade2009complexity}\nop{reference}, Fisher Information matrix~\cite{bulso2019complexity}, and the razor of model~\cite{balasubramanian1997statistical}. Here, the razor of model is a theoretical index of the complexity of a parametric family of models comparing to the true distribution.
However, deep learning models are dramatically different from those classical machine learning models discussed decades ago~\cite{montufar2014number}. The complexity analysis of classical machine learning models cannot be directly applied or straightforwardly extended to deep models. 

Recently, model complexity in deep learning has attracted more and more attention~\cite{bengio2011expressive,liang2017fisher,montufar2014number,neyshabur2014search,novak2018sensitivity,raghu2017expressive}. However, to the best of our knowledge, there is no existing survey on model complexity in deep learning. The lack of survey on this emerging and important subject motivates us to conduct this survey of the latest studies.  In this article, we use the terms ``deep learning model'' and ``deep neural network'' interchangeably.

\nop{
To investigate model complexity, we first identify the factors that affect the model complexity of deep neural networks. According to model components, a deep learning model can be divided into two parts: the static structure and the dynamic parameters~\cite{bonaccorso2017machine}. Accordingly, we categorize the factors affecting model complexity into the following four aspects.

\begin{itemize}
\item Model framework -- what kinds of models are used;
\item Model size -- the size of the model, such as width, depth and number of neurons;
\item Otimization process -- how the model is trained, such as algorithms and hyperparameters; and
\item Data complexity -- how complex the data used for training are.
\end{itemize}

A deep learning model can be regarded as a container of knowledge learned from data.  To some extent, measuring the complexity of a deep learning model is to estimate the capacity of such a container.  Therefore, two pivotal questions stand out. 

The first question is about \emph{the expressive capacity of deep models}, which investigates the power in expressing complex functions and problems~\cite{bianchini2014complexity,montufar2014number}. To some extent, this can be regarded as measuring the upper bound of deep neural networks as containers of knowledge. 
One line of work on the expressive capacity explores how the four factors listed above affect the expressive capacity of a deep model.
Among them, the effects of model depth and layer width on expressive power are the two most intensively discussed issues~\cite{lu2017expressive,montufar2014number,mhaskar2016learning,kuurkova2018constructive}. Another important direction to explore expressive capacity is to investigate the functional space that can be expressed by neural networks with certain specific structures~\cite{kileel2019expressive,arora2016understanding,khrulkov2017expressive}.

The second question is about \emph{the effective model complexity}, which investigates the practical, effective complexity of deep models with specific parameter values~\cite{raghu2017expressive,novak2018sensitivity,hu2020lann}. This is concerned about how much knowledge a specific deep neural network (possibly trained by a specific training data set) holds. The effective complexity is affected by not only model framework and model size, but also by the optimization process and data complexity, which are captured by parameter values. Designing a reasonable and sufficiently fine-grained measurement metric is an important research direction for effective model complexity, such as estimating the number of linear regions~\cite{raghu2017expressive,novak2018sensitivity} or approximating deep models~\cite{hu2020lann} to represent effective complexity. Interestingly, it is found that the actual effective complexity of a deep model may be far below the upper bound of the structure's expressive capacity~\cite{ba2014deep,hanin2019complexity}. This finding is important to explore the effectiveness of deep learning~\cite{ba2014deep}.  

To demonstrate the validity of model complexity, we discuss the applications of model complexity in generalization, optimization, model selection, and model improvement. 
Especially for generalization, from the perspective of statistical learning theory, the expressive power of a model is usually used to bound the generalization error of the model~\cite{mohri2018foundations}. Besides, several recent studies~\cite{neyshabur2017exploring,kawaguchi2017generalization,zhang2016understanding} suggest that over-parameterized deep models always exhibit good generalization performance. This finding is contrary to the belief in classic machine learning complexity~\cite{novak2018sensitivity, rasmussen2001occam}. To investigate this phenomenon in deep learning and further help understand generalization, Neyshabur~\textit{et al.}~\cite{neyshabur2017exploring} propose that a model complexity measure should have to ensure generalization capability and provide a theoretical guarantee. Subsequently, several model complexity measures are proposed to investigate generalization capability, such as the model sensitivity~\cite{novak2018sensitivity} and Fisher-Rao norm~\cite{liang2017fisher}. 
Our survey shows that the research on the complexity of deep learning models is still in the early stage. One reason is that deep learning models remain black-boxes to a large extent. In other words, we cannot fully understand yet the complex nonlinear transformations inside the boxes of sophisticated deep models.

\todo{To be revised later.}
The rest of this article is organized as follows. 
In Section~\ref{sec:concept}, we introduce the concepts of deep learning model complexity and analyze the influence factors, dimensions and properties of model complexity. 
In Section~\ref{sec:capacity}, we overview the recent studies focusing on expressive capacity of deep learning models. 
In Section~\ref{sec:effective}, we overview the latest studies on effective model complexity. 
In Section~\ref{sec:utility}, we discuss the utilities of model complexity. 
Section~\ref{sec:historyfuture} includes related works and open problems. 
Section~\ref{sec:conclusion} concludes the article. 
}


\label{sec:concept}

\nop{
In this section, we first introduce the concept of model complexity of deep neural networks. Then, we summarize the factors affecting model complexity. Third, we introduce types of complexity that have been studied intensively, the expressive capacity and the effective complexity. Finally, we identify two angles in studying model complexity, that is, model-specific versus cross-model approaches and measure-based versus reduction-based approaches. }


There are prolific studies on complexity of classical machine learning models decades ago, which are summarized in excellent surveys~\cite{buhrman2002complexity,bulso2019complexity,lim2000comparison,spiegelhalter2002bayesian}. \nop{In this section,}In the following we very briefly review the complexity of several typical models, including decision tree, logistic regression and bayesian network models. We also discuss the differences between the model complexity of deep neural networks and that of those models.

There are relatively well accepted standard measurements for complexity of decision trees. The complexity of a decision tree can be represented by the number of leaf nodes~\cite{bohanec1994trading,lim2000comparison} or the depth of the tree~\cite{buhrman2002complexity,yao1997decision}. 
Since a decision tree is constructed by recursively splitting the input space and proposing a local simple model for each resulting region~\cite{murphy2012machine}, the number of leaf nodes in principle uses the number of resulting regions to represent the complexity of the tree.   The depth of a decision tree as the complexity measure quantifies in the worst case the number of queries needed to make a classification~\cite{buhrman2002complexity}. 
A series of studies~\cite{bohanec1994trading,lim2000comparison} investigate tree optimization based on the accuracy-complexity tradeoff. 
Furthermore, Buhrman and De Wolf~\cite{buhrman2002complexity} associate the complexity of decision trees with the complexity of functions represented by certificate complexity~\cite{arora2009computational}\nop{reference?}, sensitivity complexity~\cite{cook1986upper} and approximating polynomials~\cite{nisan1994degree}\nop{reference}, and use these complexity measures of functions to bound the complexity of decision trees.

Logistic regression is the foundation of a large number of parameterized models~\cite{bulso2019complexity,kakade2009complexity}.
In the early 1990s, the degree of effective freedom is proposed as a complexity measure for linear models and penalized linear models, which is represented by Vapnik-Chervoneniks (VC) dimension~\cite{cherkassky1999model}. 
Besides, Rademacher complexity and Gaussian complexity~\cite{bartlett2002model,bartlett2002rademacher} are also used to measure model complexity of logistic regression models~\cite{kakade2009complexity}. Comparing to VC dimension, Rademacher complexity takes data distribution into consideration and therefore reflects finer-grained model complexity. 
Later, Bulso~\textit{et~al.}~\cite{bulso2019complexity} and Balasubramanian~\cite{balasubramanian1997statistical} suggest that model complexity of logistic regression models is related to the number of distinguishable distributions that can be represented by the models. Bulso~\textit{et~al.}~\cite{bulso2019complexity} define a complexity measure of logistic regression models based on the determinant of the Fisher Information matrix. 

Spiegelhater~\textit{et~al.}~\cite{spiegelhalter2002bayesian} systematically investigate the model complexity of Bayesian hierarchical models. A unique challenge in measuring the complexity of Bayesian hierarchical models is that the number of model parameters is not clearly defined. Spiegelhater~\textit{et~al.}~\cite{spiegelhalter2002bayesian} define a model complexity measure by the number of effective parameters. Using the information theoretic argument, they show that this complexity measure can be estimated by the difference between the posterior mean of the deviance and the deviance at the posterior estimates of the parameters of interest, and is approximately the trace of the product of Fisher's information~\cite{frieden2004science} and the posterior covariance matrix. In addition, they suggest that adding such a complexity measure to the posterior mean deviance gives a deviance information criterion for comparing models. 

Complexity measures are often model specific. The definition of model complexity largely depends on model structures. Different model frameworks often call for different definitions of complexity measures according to their structural characteristics. The complexity measures of different model frameworks usually cannot be directly compared.

Deep learning models are structurally different from traditional machine learning models and have dramatically more parameters. Deep learning models are always substantially more complex than traditional models. Therefore, the previous methods of model complexity of traditional machine learning models cannot be directly applied to deep learning models to obtain valid complexity measures. For example, measuring the complexity of a decision tree by the depth of the tree~\cite{buhrman2002complexity,yao1997decision} and the number of leaf nodes~\cite{bohanec1994trading,lim2000comparison} is obviously not applicable to deep learning models. Measuring model complexity by the number of trainable parameters~\cite{kakade2009complexity} has a very limited effect on deep learning models since deep learning models are often over-parameterized.


The rest of this survey is organized as follows. 
In Section~\ref{sec:basics}, we briefly introduce deep learning model complexity. 
In Section~\ref{sec:capacity}, we review the existing studies on the expressive capacity of deep learning models. 
In Section~\ref{sec:effective}, we survey the existing studies on the effective complexity of deep learning models.
In Section~\ref{sec:utility}, we discuss the applications of deep learning model complexity. 
In Section~\ref{sec:historyfuture}, we conclude this survey and discuss some future directions. 

\section{Deep Learning Model Complexity}
\label{sec:basics}

In this section, we first divide deep model complexity into two categories, expressive capacity and effective model complexity.  Then, we discuss a series of important factors of deep learning model complexity, and group the representative existing studies accordingly.

\subsection{What is Deep Learning Model Complexity?}

The term ``model complexity'' may refer to two different meanings in deep learning. 
First, model complexity may refer to capacity of deep models in expressing or approximating complicated distribution functions~\cite{bengio2011expressive}. 
Second, it may describe how complicated the distribution functions are with some parameterized deep models~\cite{raghu2017expressive}.  These two meanings are captured by the notions of model expressive capacity and model effective complexity, respectively.

 
\textbf{Expressive capacity}, also known as representation capacity, expressive power, and complexity capacity~\cite{poggio2017theory,liang2017fisher}, captures the capacity of deep learning models in approximating complex problems. Informally, the expressive capacity describes the upper bound of the complexity of any model in a family of models.

This notion is consistent with the description of hypothesis space complexity~\cite{mohri2018foundations,vapnik2013nature}. The hypothesis space is a family of hypotheses, such as the family of all neural networks with a fixed model structure. Considering the hypothesis space represented by a fixed model structure, the model expressive capacity is also the hypothesis space complexity. 
In statistical learning theory, the complexity of an infinite hypothesis space is represented by its expressive power, that is, the richness of the family of hypothesises~\cite{mohri2018foundations}. A notion to capture hypothesis space complexity is Rademacher complexity~\cite{bartlett2002rademacher}, which measures the degree to which a hypothesis space can fit random noise. Another notion is VC dimension~\cite{cherkassky1999model}, which reflects the size of the largest set that can be shattered by the hypothesis space. 
Exploring expressive capacity helps to obtain the guarantee of learnability of deep models and derive generalization bounds~\cite{mohri2018foundations}. 

\textbf{Effective model complexity}, also known as practical complexity, practical expressivity, and usable capacity~\cite{hanin2019complexity,novak2018sensitivity}, reflects the complexity of the functions represented by deep models with specific parameterization~\cite{raghu2017expressive}. The effective model complexity is for a model with parameters fixed. The study of effective model complexity helps the exploration of various aspects of deep models, such as understanding the optimization algorithms~\cite{novak2018sensitivity}, improving model selection strategies~\cite{myung2000importance} and improving model compression~\cite{cheng2017survey}.

Expressive capacity and effective model complexity are closely related but are two different concepts. Expressive capacity describes the expressive power of the hypothesis space of a deep model. Effective model complexity explores the complexity of a specific hypothesis within the hypothesis space.  Let us use an example to demonstrate the distinction and relationship between model expressive capacity and effective model complexity.

\begin{example}[Difference between expressive capacity and effective complexity]\rm
Consider unary polynomial function $f(x) = ax^2+bx+c$. The expressive capacity of $f(x)$ is unary quadratic. In other words, $f(x)$ cannot express a function more complicated than a unary quadratic polynomial. When assigning different values to the parameters $a$, $b$ and $c$, the corresponding effective complexity may be different. With parameters $a=0$, $b=1$ and $c=1$, for example, $f(x)=x+1$, the effective complexity becomes linear, which is obviously lower than the expressive capacity.  
\end{example}

Denote by $\mathcal{H}$ the hypothesis space of a fixed deep learning model structure, and by $h\in \mathcal{H}$ a hypothesis~(i.e., a deep learning model) in $\mathcal{H}$. The effective model complexity is the complexity of a specific hypothesis, written as $EMC(h)$. The 
expressive capacity
of the deep model, denoted by $MEC(\mathcal{H})$, can be written in the form of 
\begin{equation}\nonumber
	\sup \{EMC(h): h\in \mathcal{H}\} 
\end{equation}
This reflects the relationship between model expressive capacity and effective model complexity.

Because of the complex, over-parameterized architectures, deep learning models usually have high expressive capacity~\cite{montufar2014number,poole2016exponential}. However, a series of studies~\cite{ba2014deep,hanin2019complexity,novak2018sensitivity} find that the effective complexity of a trained deep model may be much lower than the expressive capacity.

Informally and intuitively, a deep learning model can be regarded as a ``container'' of knowledge learned from data. The same model architecture as a ``container'' may contain different amounts of knowledge by learning from different data and thus equipped with different parameters. 
The expressive capacity can be regarded as the upper bound of the amount of knowledge that a model architecture can hold. 
The effective model complexity is concerned about, for a specific model, a specific training dataset, how much knowledge it actually holds.


\subsection{Important Factors of Deep Learning Model Complexity}
\label{sec:concept:factors}

Bonaccorso~\cite{bonaccorso2017machine} points out that a deep learning model consists of a static structure part and a dynamic parametric part. The static structure part is always determined before the learning process by the model selection principle, then stays immutable once decided. The parametric part is the objective of the optimization and is determined by a learning process. Both the static and dynamic parts contribute to model complexity. We refine this division and summarize four aspects affecting model complexity, including both expressive capacity and effective complexity.

\begin{description}
	\item [\textbf{Model framework}] The choice of model framework affects model complexity. The factor of model framework includes model type (e.g.,  feedforward neural network, convolutional neural network), activation function (e.g., Sigmoid~\cite{cybenko1989approximation}, ReLU~\cite{nair2010rectified}), and others. Different model frameworks may require different complexity measure criteria and may not be directly comparable to each other. 
	
	\item [\textbf{Model size}] The size of deep model affects model complexity. Some commonly used measures of model size include the number of parameters, the number of hidden layers, the width of hidden layers, the number of filters, and the filter size. Under the same model framework, the complexities of models with different sizes can be quantified by the same complexity measure criteria and thus become comparable~\cite{kuurkova2018constructive}. 
	
	\item [\textbf{Optimization process}] The optimization process affects model complexity, including the form of objective functions, the selection of learning algorithms, and the setting of hyperparameters. 
	
	\item [\textbf{Data complexity}] The data on which a model is trained affect model complexity, too. The major factors include the data dimensionality, data distribution~\cite{mohri2018foundations}, information volume measured by Kolmogorov complexity~\cite{cano2013analysis,li2006data}, and some others.	
\end{description}

Among these four aspects, model framework and model size mainly affect the static structural part of a deep model, and optimization process and data complexity mainly affect the dynamic parametric part.

\nop{
\begin{figure}[t]
	\centering
	\includegraphics[width = 0.8\linewidth]{figures/complexity_framework}
	\caption{A deep neural network consists of a static structure part and a dynamic parametric part. Among the four aspects discussed in Section~\ref{sec:concept:factors}, model framework and model size affect the static structure part, and learning process and data complexity affect the dynamic parametric part. \todo{Add ``Static structure part'' and ``Dynamic parametric part'' to the left of the two boxes.  ``Learning Procedure'' should be ``Optimization Process''.}}
	\label{fig:complexity_framework}
\end{figure}
}

\begin{table*}[t]
	\begin{center}
		\caption{Summarize the aspects affecting the expressive capacity and effective complexity, respectively. 
		}
		\label{tab:twoprobproperty}
		\resizebox{\linewidth}{!}{
			\begin{tabular}{c|c|cccc}
				\hline
				& & Model & Model & Learning & Data \\
				& & Framework & Size & Process & Complexity \\
				\hline
				\multirow{4}*{Expressive Capacity} &  Depth Efficiency &&\checkmark && \\
				&  Width Efficiency & & \checkmark && \\
				&Expressible Functional Space &\checkmark &\checkmark &\checkmark&\checkmark \\
				& VC Dimension and Rademacher Complexity & \checkmark & \checkmark & & \checkmark \\
				\hline
				\multirow{2}*{Effective Complexity} & Effective Complexity Measure &\checkmark &\checkmark&\checkmark&\checkmark \\
				& High-capacity Low-reality Phenomenon & \checkmark &\checkmark & \checkmark & \checkmark  \\
				\hline
			\end{tabular}
		}
	\end{center}
\end{table*}

The effect of these four aspects on expressive capacity can be understood through the influence on the hypothesis space. 
A selected model framework corresponds to a hypothesis space $\mathcal{H}$. Each hypothesis $h\in \mathcal{H}$ represents a model with the given framework. Once the model size is determined, the hypothesis space shrinks to a subset of $\mathcal{H}$.
For example, suppose we set up two models with depth $l_1$ and $l_2$ ($l_1 \leq l_2$), respectively, and both with width $m$, the corresponding hypothesis spaces are $\mathcal{H}_1$ and $\mathcal{H}_2$, respectively. We have $\mathcal{H}_1 \subset \mathcal{H}_2$. This is because, for each $h \in \mathcal{H}_1$, where $h$ is a hypothesis and thus a deep network model in this context, there exists $h'\in \mathcal{H}_2$ whose first $l_1$ layers are identical to those in $h$ and the subsequent layers are identical mappings. It is easy to show that the expressive capacity of $\mathcal{H}_1$ will not exceed the expressive capacity of $\mathcal{H}_2$.
Recently, model framework, model size and their effects on expressive capacity of deep models have been well explored~\cite{bengio2011expressive,khrulkov2017expressive,lu2017expressive,raghu2017expressive}.

The choice of data distribution and optimization algorithm will further reduce the scope of the hypothesis space, thereby affecting the expression capacity.
\nop{Data distribution of training data also effects the expressive power.}
For example, Rademacher complexity is a data-dependent expressive capacity measure taking into account the effect of data distribution~\cite{bartlett2002rademacher,mohri2018foundations}. However, to the best of our knowledge, the effect of data complexity and optimization process on expressive capacity of deep learning models is still rarely explored.

These four aspects also affect the effective model complexity. In general, model framework and model size decide the available range of effective model complexity. 
\nop{The effective complexity of the model is selected from this range by optimization process and training data.}
The effective complexity of a model with fixed parameters is a value in this range selected by optimization process and training data. The optimization process affects the effective complexity, for example, adding $L^1$ regularization constrains the degrees of freedom in a deep neural network, and thus constrains the effective model complexity~\cite{hu2020lann,kakade2009complexity}. 
The training data affects the effecive complexity, for example, using the same model and the same optimization process, the effective complexity of a model trained on linearly classifiable data is much lower than that trained on the ImageNet~\cite{deng2009imagenet,li2006data}.
Specifically, the effect of training data complexity and optimization process on effective complexity are reflected in the values of model parameters.

In Table~\ref{tab:twoprobproperty} we list the corresponding aspects affecting expressive capacity and effective complexity, respectively. 
In Table~\ref{tab:overview} we list some representative studies on the two major problems, expressive capacity and effective complexity.

\begin{sidewaystable}[!ph]
	\begin{center}
		\caption{Overview of a collection of studies on model complexity of deep neural networks.}
		\label{tab:overview}
		\renewcommand{\arraystretch}{1.5}
		\resizebox{0.95\linewidth}{!}{
		\begin{tabular}{|c|P{3.5cm}||c|c|c|c|c|P{6.5cm}|P{6.5cm}|}
			\hline
			& & References & \rotatebox{70}{Model-specific} & \rotatebox{70}{Cross-model} & \rotatebox{70}{Measure-based} & \rotatebox{70}{Reduction-based} & Objective Model & Measure Metric \\ 
			\hline\hline
			\multirow{15}{*}{\rotatebox{90}{Expressive Capacity}} & \multirow{6}{*}{Depth Efficiency} &  \cite{bengio2011expressive,delalleau2011shallow} & $\bullet$ &  &  & $\bullet$ & SPN & -  \\ \cline{3-9} 
			&  & \cite{mhaskar2016learning} & $\bullet$ &  &  & $\bullet$ & Binary tree hierarchical network, deep Gaussian network & - \\ \cline{3-9} 
			&  & \cite{kuurkova2018constructive} & $\bullet$ &  &  & $\bullet$ & FCNN with $\sigma(z)=\{Signum, Heaviside\}$ & -\nop{Variational norm} \\ \cline{3-9} 
			&  &  \cite{montufar2014number} & $\bullet$ &  & $\bullet$ &  & FCNN with $\sigma(z)=\{ReLU, Maxout\}$ & \#linear regions \\ \cline{3-9} 
			&  &  \cite{serra2018bounding} & $\bullet$ &  & $\bullet$ &  & FCNN with $\sigma(z)=\{ReLU, Maxout\}$ & \#linear regions \\ \cline{3-9} 
			&  & \cite{bianchini2014complexitydepth,bianchini2014complexity} &  & $\bullet$ & $\bullet$ &  & FCNN & Sum of Betti number \\ \cline{2-9} 
			
			& Width Efficiency & \cite{lu2017expressive} & $\bullet$ &  &  & $\bullet$ & FCNN with $\sigma(z)=\{ReLU\}$ & - \\ \cline{2-9} 
			
			& \multirow{4}{*}{\shortstack{Expressible Functional\\ Space}} & 
			\cite{arora2016understanding} & $\bullet$ &  &  & $\bullet$ & FCNN with $\sigma(z)=\{ReLU\}$ & - \\ \cline{3-9} 
			&  & \cite{guhring2019complexity} & $\bullet$ &  &  & $\bullet$ & FCNN with $\sigma(z)=\{ReLU\}$ & - \\ \cline{3-9} 
			&  & \cite{kileel2019expressive} & $\bullet$ &  & $\bullet$ &  & FCNN with $\sigma(z)=\{\sigma(z)=z^r\}$ & Dimension of functional variety \\ \cline{3-9}
			& & \cite{khrulkov2017expressive} &  & $\bullet$ & & $\bullet$ & FCNN, CNN, RNN & Rank of tensor decomposition \\ \cline{2-9} 
			
			& \multirow{5}{*}{\shortstack{VC Dimension and\\Rademacher Complexity}} & \cite{maass1994neural} & $\bullet$ & & $\bullet$ & & FCNN with linear threshold gates & VC dimension \\ \cline{3 - 9}
			& & \cite{bartlett1998almost} & $\bullet$ & & $\bullet$ & & FCNN with $\sigma=\{piecewise\ polynomial\}$ & VC dimension \\ \cline{3 - 9}
			& & \cite{bartlett2019nearly} & $\bullet$ & & $\bullet$ & & Networks with $\sigma=\{piecewise\ linear\}$ &  VC dimension \\ \cline{3 - 9}
			& & \cite{bartlett2017spectrally} & $\bullet$ & & $\bullet$ & & Networks with $\sigma=\{ReLU\}$ & Rademacher complexity \\ \cline{3 - 9}
			& & \cite{neyshabur2018role} &  $\bullet$ & & $\bullet$ & & Two-layer FCNN with $\sigma=\{ReLU\}$ & Rademacher complexity \\ \cline{3 - 9}
			\hline
			\multirow{7}{*}{\rotatebox{90}{Effective Complexity}} & \multirow{5}{*}{\shortstack{Effective Complexity \\Measure}} &  \cite{raghu2017expressive} & $\bullet$ &  & $\bullet$ &  & Networks with $\sigma=\{piecewise\ linear\}$ & \#linear regions, trajectory length \\ \cline{3-9} 
			&  & \cite{novak2018sensitivity} & $\bullet$ &  & $\bullet$ &  & Networks with $\sigma=\{piecewise\ linear\}$ & Jacobian norm, trajectory length \\ \cline{3-9} 
			&  & \cite{hu2020lann} & $\bullet$ &  & $\bullet$ &  & FCNN with $\sigma(z)=\{curve\ activation\}$ & \#approximated linear regions \\ \cline{3-9} 
			&  & \cite{liang2017fisher} & $\bullet$ &  & $\bullet$ &  & FCNN with $\sigma(z)=\sigma'(z)z$ & Fisher-Rao norm \\ \cline{3-9} 
			&  & \cite{nakkiran2019deep} & & $\bullet$  & $\bullet$ &  & Any deep models & max \#samples to zero training error\\ \cline{2-9} 
			& High-Capacity Low-Reality Phenomenon &\cite{hanin2019complexity} & $\bullet$ &  & $\bullet$ &  & FCNN with $\sigma(z)=\{ReLU\}$ & \#linear regions, volumn of region boundaries \\ \hline
		\end{tabular}		
		}
	\end{center}
\end{sidewaystable}

\subsection{Existing Studies on Deep Learning Model Complexity: An Overview}
\label{sec:category}

The literature on deep model complexity can be categorized from two different angles. The first angle focuses on whether an approach is model-specific or cross-model. 
The second angle focuses on whether an approach develops an explicit complexity measure. 
These two angles are applicable to  both expressive capacity and effective complexity. 

\subsubsection{Model-Specific versus Cross-Model}
\label{sec:category-1}

Based on whether an approach focuses on one type of models or crossing multiple types of models, the existing studies on model complexity can be divided into two groups, model-specific approaches and cross-model approaches.

A \textbf{model-specific} approach focuses on a certain type of models, and explores complexity based on structural characteristics. For example,  Bianchini~\textit{et~al.}~\cite{bianchini2014complexitydepth,bianchini2014complexity} and Hanin~\textit{et~al.}~\cite{hanin2019complexity} study model complexity of fully-connected feedforward neural networks (FCNNs for short), Bengio and Delalleau~\cite{bengio2011expressive,delalleau2011shallow} focus on model complexity of sum-product networks. Moreover, some studies further propose constraints on the activation functions to constrain the nonlinearity properties. For example, Liang~\textit{et~al.}~\cite{liang2017fisher} assume that activation functions $\sigma(\cdot)$ satisfy $\sigma(z) = \sigma'(z)z$.

An approach is \textbf{cross-model} when it covers multiple types of models rather than one specific type of models,  and thus can be applied to compare two or more different types of models. For example, Khrulkov~\textit{et~al.}~\cite{khrulkov2017expressive} compare the complexity of general recurrent neural networks (RNNs for short), convolutional neural networks (CNNs for short) and shallow FCNNs by building connections among these network architectures and tensor decompositions. 

\subsubsection{Measure-Based versus Reduction-Based}
\label{sec:category-2}

According to whether an explicit measure is designed, the state-of-the-art model complexity approaches can be divided into two groups, measure-based approaches and reduction-based approaches.

A deep model complexity study is \textbf{measure-based} if it defines an appropriate quantitative representation of model complexity. For example, the number of linear regions is used to represent the complexity of FCNNs with piecewise linear activation functions~\cite{hanin2019complexity,montufar2014number,novak2018sensitivity,raghu2017expressive}.

A complexity approach is \textbf{reduction-based} if it investigates a model complexity problem by reducing deep networks to some known problems and functions, and does not define any explicit complexity measure. For example, Arora~\textit{et~al.}~\cite{arora2016understanding} build connections between deep networks with ReLU activation functions and Lebesgue spaces. Khrulkov~\textit{et~al.}~\cite{khrulkov2017expressive} connect deep neural networks with several types of tensor decompositions.

\nop{
In addition to the above two angles, we suggest that a key desideratum of model complexity study is the comparability. That is, a study of model complexity is expected to provide distinctions and comparisons between models, and thus can answer questions like, given two deep neural networks with different structures, which one has better expressive power; given two similar networks but with different training algorithms, which one ensures better generalization.
}

\section{Expressive Capacity of Deep Learning Models}
\label{sec:capacity}

Expressive capacity of deep models is also known as representation capacity, expressive power, and complexity capacity~\cite{liang2017fisher,poggio2017theory}. It describes how well a deep learning model can approximate complex problems. Informally, the expressive capacity describes the upper bound of the complexity in a parametric family of models. 

Expressive capacity of deep learning models has been mainly explored from four aspects.
\begin{itemize}
	\item \textbf{Depth efficiency} analyzes how deep learning models gain performance (e.g., accuracy) from the depth of architectures.
	\item \textbf{Width efficiency} analyzes how the widths of layers in deep learning models affect model expressive capacity.
	\item \textbf{Expressible functional space} investigates the functions that can be expressed by a deep model with a specific framework and specified size using different parameters. 
	\item \textbf{VC Dimension and Rademacher Complexity} are two classic measures of expressive capacity in machine learning.
\end{itemize}

In this section, we review these four groups of studies. 


\subsection{Depth Efficiency}
A series of recent studies demonstrate that deep architectures significantly outperform shallow ones~\cite{bengio2011expressive,mhaskar2016learning}. Depth efficiency~\cite{lu2017expressive}, which is the effectiveness of depth of deep models, has attracted a lot of interest. Specifically, the studies on depth efficiency analyze why deep architectures can obtain good performance and measures the effects of model depth on expressive capacity.  We divide the studies of depth efficiency into two subcategories: model reduction methods and expressive capacity measures.

\subsubsection{Model Reduction}

One way to study the expressive capacity of deep learning models is to reduce deep learning models to understandable problems and functions for analysis.

To investigate depth efficiency, one intuitive idea is to compare the representation efficiency between deep networks and shallow ones.
Bengio and Delalleau~\cite{bengio2011expressive,delalleau2011shallow} investigate the depth efficiency problem on deep sum-product networks (SPN for short). An SPN consists of neurons computing either products or weighted sums of their inputs. 
They consider two families of functions built from deep sum-product networks. The first family of functions is 
$F= \cup_{n \geq 4}F_n$,
where $F_n$ is the family of functions represented by deep SPNs with $n=2^k$ inputs and depth $k$. The second family of functions is $G = \cup_{n\geq 2, i\geq 0} G_{i,n}$,
where $G_{i,n}$ is the family of functions represented by SPNs with $n$ inputs and depth $2i+1$. 

Then, Bengio and Delalleau establish the lower bounds on the number of hidden neurons required by a shallow sum-product network to represent those families of functions. To approach a function $f \in F$, a one-hidden-layer shallow SPN needs at least $2^{\sqrt{n}-1}$ hidden neurons and at least $2^{\sqrt{n}-1}$ product neurons. Similarly, to approach a function $g \in G$, a one-hidden-layer shallow SPN needs at least $(n-1)^i$ hidden neurons. 
The comparison of deep and shallow sum-product networks representing the same function\nop{taken from $F$ and $G$} indicates that, to represent the same functions, the number of neurons in a shallow network has to grow exponentially but only a linear growth is needed for deep networks. 

Mhaskar~\textit{et al.}~\cite{mhaskar2016learning} study functions representable by hierarchical binary tree networks, comparing with shallow networks.  Fig.~\ref{fig:mhaskar2016learning} demonstrates shallow networks and hierarchical binary tree networks. 
Denote by $S_m$ the class of shallow networks with $m$ neurons, in the form of
\begin{equation}\nonumber
\label{eq:mhaskar2016:2}
x \rightarrow \sum_{k=1}^{m} a_k\sigma (w_k\cdot x+b_k)
\end{equation}
where $w_k \in \mathbb{R}^d$, $b_k, a_k \in \mathbb{R}$ are the parameters for the $k$-th hidden neuron, $\sigma$ is the activation function. 
Denote by $W^{NN}_{r,d}$ the class of functions with continuous partial derivatives of order $\leq r$ with certain assumptions (see~\cite{mhaskar2016learning} for detail).
Correspondingly, a hierarchical binary tree network is a network with the structure 
\begin{equation}
\label{eq:mhaskar2016:1}
\begin{split}
f(x_1, \ldots, x_d) = h_{l1}(& h_{(l-1), 1}(\ldots (h_{11}(x_1, x_2), h_{12}(x_3, x_4), \ldots), \\
& h_{(l-1), 2} (\ldots))
\end{split}
\end{equation}
where each $h_{ij}$ is in $S_m$, $l$ is the depth of the network, $h_{l1}$ corresponds to the root node of the tree structure. See Fig.~\ref{fig:mhaskar2016learning:tree} as an example of the hierarchical binary tree with depth$=3$. Denote by $D_m$ the class of hierarchical binary tree networks. 
Let $W^{NN}_{H, r, d}$ be the class of functions with the structure of Eq.~(\ref{eq:mhaskar2016:1}), where each $h_{ij}$ is in $W^{NN}_{r,2}$.
Define by $\inf_{P\in V} ||f - P||$ the approximation degree of $V$ to function $f$, where $V = \{S_m, D_m\}$.

\begin{figure}[t]
	\centering
	\subfigure[]{
		\includegraphics[width = 0.46\linewidth]{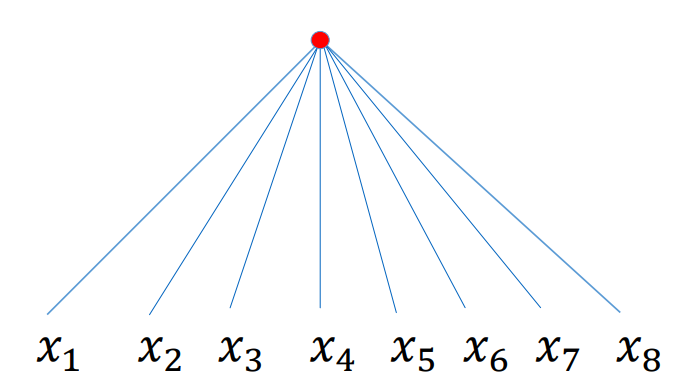}
		\label{fig:mhaskar2016learning:shallow}
	}
	\subfigure[]{
		\includegraphics[width=0.42\linewidth]{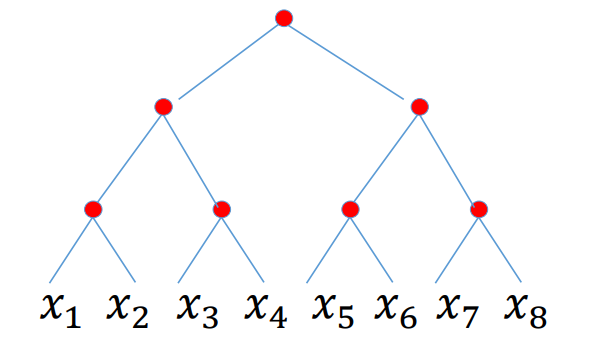}
		\label{fig:mhaskar2016learning:tree}
	}
	\caption{8 input dimensions fed into a shallow network (a) and a hierarchical binary tree network (b), studied by Mhaskar~\textit{et~al.}~\cite{mhaskar2016learning}.}
	\label{fig:mhaskar2016learning}
\end{figure}

Mhaskar~\textit{et al.}~\cite{mhaskar2016learning} demonstrate that, to approximate a function $f\in W^{NN}_{r,d}$ to approximation degree $\epsilon$, a shallow network in $S_m$ requires $O(\epsilon^{-d/r})$ trainable parameters. Meanwhile, to approximate a function $f \in W^{NN}_{H, r, d}$ to the same approximation degree, a network in $D_m$ requires only $O(\epsilon^{-2/r})$ trainable parameters. 
Then, Mhaskar~\textit{et al.}~\cite{mhaskar2016learning} compare shallow Gaussian networks with hierarchical binary tree structures (Eq.~(\ref{eq:mhaskar2016:1})) where each $h_{ij}$ computes a shallow Gaussian network, and obtain a similar conclusion. They demonstrate that functions with a designed compositional structure can be approximated by both deep and shallow networks to the same degree of approximation. However, the numbers of parameters of deep networks are much less than those of shallow networks.

\nop{To approximate a structure-corresponding $f$ to $\epsilon$ accuracy, a shallow Gaussian network requires $O(\epsilon^{-2L/r})$ trainable parameters while a deep one requires only $O(L\epsilon^{-4/r})$ trainable parameters. Here $L$ denotes the model depth.
where shallow Gaussian network is with form of 
\begin{equation}
	\sum_{k=1}^{n}a_k \exp (-|\bm{x}-\bm{x}_k|^2)
\end{equation}
, $x\in \mathbb{R}^d$ and deep Gaussian the compositional function of shallow Gaussian. To approximate a structure-corresponding $f$ to accuracy of $\epsilon$, a shallow Gaussian network requires $O(\epsilon^{-2L/r})$ trainable parameters while deep one requires only $O(L\epsilon^{-4/r})$ trainable parameters. }

Arora~\textit{et al.}~\cite{arora2016understanding} investigate the importance of depth on deep neural networks with ReLU activation function. 
First, they investigate neural networks with one-dimensional input and one-dimensional output. They prove that given any natural numbers $k \geq 1$ and $w \geq 2$, there exists a family of functions that can be represented by a ReLU neural network with $k$ hidden layers each of width $w$. However, to represent this family of functions, a network with $k' < k$ hidden layers requires at least $\frac{1}{2}k'w^{\frac{k}{k'}}-1$ hidden neurons. 
Then, they investigate ReLU neural networks with $d$ input dimensions. They prove that,  given natural numbers $k, m,d \geq 1$ and $w \geq 2$, 
there exists a family of $\mathbb{R}^d \rightarrow \mathbb{R}$ functions that can be represented by a ReLU network with $k+1$ hidden layers and $2m+wk$ neurons. This family of functions is constructed using the zonotope theory from the polyhedral theory~\cite{ziegler1993lectures}. However, to represent this family of functions, the minimum number of hidden neurons that a ReLU neural network with $k'\leq k$ hidden layers requires is
$$
\max\{\frac{1}{2}(k'w^{\frac{k}{k'd}})(m-1)^{(1-\frac{1}{d})\frac{1}{k'}} - 1, k'(\frac{w^{\frac{k}{k'}}}{d^{\frac{1}{k'}}})\}.
$$ 

\nop{\begin{figure}
	\centering
	\includegraphics[width=0.3\linewidth]{figures/capacity/kurkova2016constructive_sh}
	\caption{A $2^4\times 2^4$ Sylvester-Hadamard matrix~\cite{kuurkova2018constructive}. }
	\label{fig:mhaskar2016}
\end{figure}}

To investigate when deep neural networks are provably more efficient than shallow ones, Kuurkova~\cite{kuurkova2018constructive} analyzes the limitation of expressive capacity of shallow neural networks with Signum activation function. The Signum activation function is defined as 
\begin{equation}\nonumber
sgn(z) = \left\{ \begin{array}{rcl}
-1 &\quad \mbox{for} & z<0 \\ 
1 &\quad \mbox{for} & z \geq 0
\end{array}\right .
\end{equation}
Kuurkova proves that there exist functions that cannot be $L^1$ sparsely represented by one-hidden-layer Signum neural networks, which have a limited number of neurons and a limited sum of absolute values of output weights (i.e., $L^1$-norm). Such functions should be nearly orthogonal to any function $f$ from the class of Signum perceptrons
$
	\{sgn(vx+b): X \rightarrow \{-1, 1\} | v \in \mathbb{R}^d, b \in \mathbb{R}\}
$. The functions generated by Hadamard matrices are such examples. A Hadamard matrix of order $n$ is a $n\times n$ matrix with entries in $\{-1, 1\}$ such that any two distinct rows or columns are orthogonal. Kuurkova~\cite{kuurkova2018constructive} proves that the functions induced by $n\times n$ Hadamard matrices cannot be computed by shallow Signum networks with both the number of hidden neurons and the sum of absolute values of output weights smaller than $\frac{\sqrt{n}}{\lceil \log_2 n\rceil}$. 


To further illustrate the limitation of shallow networks, Kuurkova~\cite{kuurkova2018constructive} compares the representation capability of one and two hidden layer networks with Heaviside activation function. The Heaviside activation function is defined as
\begin{equation}
\sigma(z) = \left \{ \begin{array}{rcl}
0 & \quad \mbox{for} & z<0 \\
1 & \quad \mbox{for} & z \geq 0
\end{array}\right .
\label{eq:heaviside}
\end{equation}
Let $S(k)$ be a $2^k\times 2^k$ Sylvester-Hadamard matrix, which is constructed by starting from $S(2) = \left ( \begin{array}{rcl} 1 & 1 \\ 1 & -1 \end{array}\right)$ and then recursively iterating $S(l+1) = S(2)\otimes S(l)$. Kuurkova~\cite{kuurkova2018constructive} shows that, to represent a function induced by $S(k)$, a two-hidden-layer Heaviside network requires $k$ neurons in each hidden layer. However, to represent such a function, a one-hidden-layer Heaviside network requires at least $\frac{2^k}{k}$ hidden neurons, or the absolute value of some output weights is no less than $\frac{2^k}{k}$.

In summary, the model reduction approaches reduce neural networks to some kind of functions~\cite{arora2016understanding,bengio2011expressive,kuurkova2018constructive,mhaskar2016learning},  and investigate the effects of model depth on the capacity to express function families.

\subsubsection{Expressive Capacity Measures}

To investigate depth efficiency, another idea is to develop an appropriate measure of expressive capacity and study how expressive capacity changes when the depth and layer width of a model increase. 


Montufar~\textit{et al.}~\cite{montufar2014number} focus on fully-connected feed forward neural networks (FCNNs) with piecewise linear activation functions (e.g., ReLU~\cite{nair2010rectified} and Maxout~\cite{goodfellow2013maxout}), and propose to use the number of linear regions as a representation of model complexity (Fig.~\ref{fig:montufar2014number}). 
The basic idea is that an FCNN with piecewise linear activation functions divides the input space into a large number of linear regions. Each linear region corresponds to a linear function. The number of linear regions can reflect the flexibility and complexity of the model. 

\begin{figure}[t]
	\centering
	\includegraphics[width=0.6\linewidth]{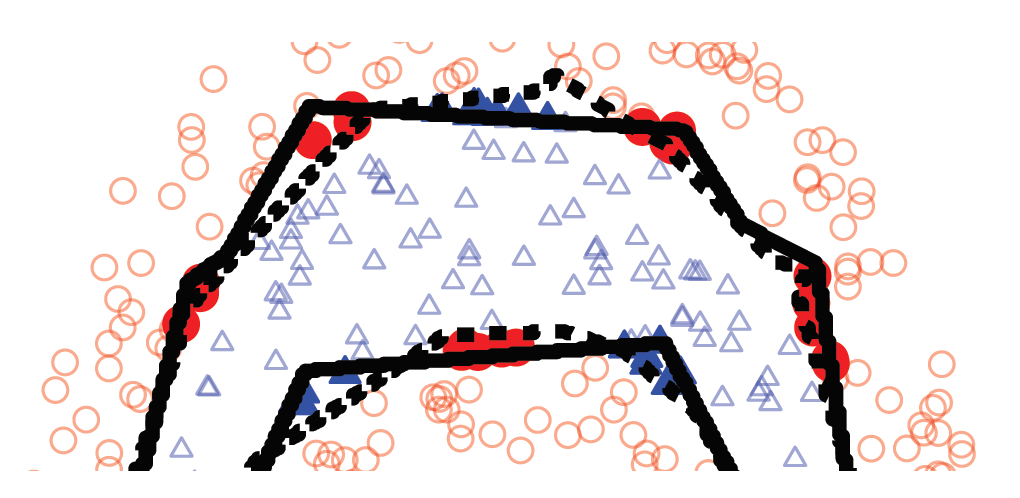}
	\caption{An example from Montufar~\textit{et al.}~\cite{montufar2014number} illustrating the advantage of deep models. A deep ReLU network (the dotted line) captures the boundary more accurately by approximating it with a larger number of linear regions than a shallow one (the solid line).}
	\label{fig:montufar2014number}
\end{figure}

Montufar~\textit{et al.}~\cite{montufar2014number} investigate FCNNs with two types of piecewise linear activation functions: ReLU~\cite{nair2010rectified} and Maxout~\cite{goodfellow2013maxout}. They prove that, under certain parameter settings, the model expressive capacity of a ReLU network $\mathcal{N}$, which is represented by the maximum number of linear regions and denoted by $MEC(\mathcal{N})$, is bounded by
\begin{equation}
	\label{eq:montufar14:lowerbound}
	MEC(\mathcal{N}) \geq (\prod_{i=1}^{l-1}\lfloor \frac{m_i}{m_0}\rfloor^{m_0})\sum_{j=0}^{m_0}\tbinom{m_l}{j}
\end{equation}
where $l$ is the number of hidden layers, $m_i$ is the width of $i$-th hidden layer, $m_0 = d$ is the dimensionality of the input. Based on this bound, they show that a ReLU network with  $l$ hidden layers and layer width $m_i \geq m_0$ is able to approximate any piecewise linear function that has $\Omega((\frac{m}{m_0})^{(l-1)m_0}m^{m_0})$ linear regions. 

Montufar~\textit{et al.}~\cite{montufar2014number} also prove that, for the rank-$k$ Maxout activation function,  the expressive capacity of a one-hidden-layer Maxout network with $m$ neurons is bounded by
$
	MEC(\mathcal{N}) \geq k^{\min(d, m)}
$
and $
	MEC(\mathcal{N}) \leq \min \{\sum_{j=1}^{d} \tbinom{k^2m}{j},\ k^m\}
$.
A rank-$k$ Maxout network, which consists of $l$ hidden layers and the width of each layer equals $m$, can compute any piecewise linear function with $\Omega(k^{l-1}k^m)$ linear regions.  In conclusion, \nop{they}the maximum number of linear regions generated by a FCNN with piecewise linear activation functions increases exponentially with model depth. 

Montufar~\textit{et al.}~\cite{montufar2014number} provide an explanation for the depth efficiency. They suggest that the intermediary layer of a deep model is able to map several pieces of the inputs into the same output. As the number of layers increases, the layer-wise compositions of the functions re-use lower-level computation exponentially. This allows deep models to compute highly complex functions, even with relatively fewer parameters. 

Serra~\textit{et al.}~\cite{serra2018bounding} improve the bounds of the maximum number of linear regions proposed by Montufar~\textit{et al.}~\cite{montufar2014number} (Eq.~(\ref{eq:montufar14:lowerbound})). 
Given a deep ReLU neural network with $l$ layers, let $m_i$ be the width of $i$-th hidden layer with $m_i \geq 3d$, $d$ is the input dimension. Serra~\textit{et al.}~\cite{serra2018bounding} prove that the maximal number of linear regions of this neural network is lower bounded by
\begin{equation}\nonumber
	MEC(\mathcal{N}) \geq (\prod_{i=1}^{l-1}(\lfloor \frac{m_i}{d}\rfloor+1)^{d})
	\sum_{j=0}^{d}\tbinom{m_l}{j}
\end{equation}
and is upper bounded by
\begin{equation}\nonumber
	MEC(\mathcal{N}) \leq \sum_{(j_1,\ldots,j_{l+1})\in J} \prod_{i=1}^{l+1} \tbinom{m_i}{j_i}
\end{equation}
where $J = \{(j_1,\ldots, j_{l+1}) \in \mathbb{Z}^{l+1}: 0 \leq j_i \leq \min\{d, m_1-j_1, \ldots, m_{i-1}-j_{i-1}, m_i\}\}$.

Bianchini and Scarselli~\cite{bianchini2014complexitydepth,bianchini2014complexity} design a topological measure of model complexity for deep neural networks. Given an FCNN with single output, denoted by $\mathcal{N}: \mathbb{R}^d \rightarrow \mathbb{R}$, they define $S=\{x \in \mathbb{R}^d | \mathcal{N}(x) \geq 0\}$ the set of instances being classified by $\mathcal{N}$ to the positive class. The model complexity of neural network $\mathcal{N}$ is measured by $B(S)$, the sum of the Betti numbers of set $S$, that is,
\begin{equation}\nonumber
	MEC(\mathcal{N}) = B(S) = \sum_{i=0}^{d-1} b_i(S)
\end{equation}
where $b_i(S)$ denotes the $i$-th Betti number that counts the number of $(i+1)$-dimensional holes in $S$. The Betti number is generally used to distinguish between spaces with different characteristics in algebraic topology~\cite{bredon2013topology}. $S$ contains instances positively classified by the network $\mathcal{N}$. Therefore, $B(S)$ can be used to investigate how $S$ is affected by the architecture of $\mathcal{N}$ and can represent the model complexity.

Bianchini and Scarselli~\cite{bianchini2014complexitydepth,bianchini2014complexity} report upper and lower bounds of $B(S)$ for a series of network architectures~(Table~\ref{tab:bianchini2014complexity}).
In particular, Bianchini and Scarselli~\cite{bianchini2014complexitydepth,bianchini2014complexity} demonstrate that $B(S)$ of a single-hidden-layer network grows polynomially with respect to the hidden layer width. That is, $B(S) \in O(m^d)$, where $m$ is the width of the hidden layer. They also prove that $B(S)$ of a deep neural network grows exponentially in the total number of hidden neurons. That is, $B(S) \in O(2^M)$, where $M$ is the total number of hidden neurons. This indicates that deep neural networks have higher expressive capacity, and therefore are able to learn more complex functions than shallow ones. 

\begin{table*}[t]
	\begin{center}
		\caption{Upper and lower bounds of $B(S)$ given by Bianchini and Scarselli~\cite{bianchini2014complexity,bianchini2014complexitydepth}, for networks with a number of $M$ hidden units, a number of $d$ inputs and a number of $l$ hidden layers.}
		\label{tab:bianchini2014complexity}
		\resizebox{\linewidth}{!}{
		\begin{tabular}{lllc}
			\hline
			\#Inputs & \#Hidden Layers & Activation Function & Bound of $B(S)$ \\
			\hline\hline
			\multicolumn{4}{c}{Upper Bounds of $B(S)$} \\
			\hline
			d & 1 & threshold & $O(M^d)$ \\
			d & 1 & arctan & $O((d+M)^{d+2})$ \\
			d & 1 & polynomial, degree $r$ & $\frac{1}{2}(2+r)(1+r)^{d-1}$ \\
			1 & 1 & arctan & $M$ \\
			d & many & arctan & $2^{M(2M-1)}O((dl+d)^{d+2M})$ \\
			d & many & tanh & $2^{M(M-1)/2}O((dl+d)^{d+M})$ \\
			d & many & polynomial, degree $r$ & $\frac{1}{2}(2+r^l)(1+r^l)^{d-1}$ \\
			\hline\hline
			\multicolumn{4}{c}{Lower Bounds of $B(S)$} \\
			\hline
			d & 1 & any sigmoid & $(\frac{M-1}{d})^d$ \\
			d & many & any sigmoid & $2^{l-1}$ \\
			d & many & polynomial, degree $r\geq 2$ & $2^{l-1}$ \\
			\hline
		\end{tabular}
		}
	\end{center}
\end{table*}

Bianchini and Scarselli~\cite{bianchini2014complexitydepth,bianchini2014complexity} suggest that the layer-wise composition mechanism of a deep model allows the model to replicate the same behavior in different regions of the input space, thus makes depth more effective than width.

\subsection{Width Efficiency}

In addition to depth efficiency, the effect of width on expressive capacity, namely width efficiency, is also worth exploring. Width efficiency analyzes how width affects the expressive capacity of deep learning models~\cite{lu2017expressive}. Width efficiency is important for fully understanding expressive capacity and helps to validate the insights obtained from depth efficiency~\cite{lu2017expressive}. 

Lu~\textit{et al.}~\cite{lu2017expressive} investigate the width efficiency of neural networks with ReLU activation function.
They extend the universal approximation theorem~\cite{barron1993universal,hornik1989multilayer} to width-bounded deep ReLU neural networks. The classical universal approximation theorem~\cite{barron1993universal,hornik1989multilayer} states that, one-hidden-layer neural networks with certain activation functions (e.g., ReLU) can approximate any continuous functions on a compact domain to any desired accuracy performance. 
Lu~\textit{et al.}~\cite{lu2017expressive} show that, for any Lebesgue-integrable function $f:\mathbb{R}^d \rightarrow \mathbb{R}$ and any $\epsilon > 0$, there exists a ReLU network $\mathcal{N}: \mathbb{R}^d \rightarrow \mathbb{R}$ \nop{with width of each hidden layer being contrained by } that can approximate $f$ to $\epsilon$ $L^1$-distance. That is,
\begin{equation}\nonumber
	\int_{\mathbb{R}^d} |f(x) - F_{\mathcal{N}}(x)|dx < \epsilon.
\end{equation}
Here $F_{\mathcal{N}}$ is the function represented by the neural network $\mathcal{N}$. The width $m_i$ of each hidden layer of $\mathcal{N}$ satisfies $m_i \leq d+4$.

Moreover, to explore the role of layer width in expressive capacity quantitatively, Lu~\textit{et al.}~\cite{lu2017expressive} raise the dual question of depth efficiency. That is, whether there exists wide, shallow ReLU networks that cannot be approximated by any narrow, deep neural network whose size is not substantially increased. 
Denote by $F_{\mathcal{A}}: \mathbb{R}^d \rightarrow \mathbb{R}$ the function represented by a ReLU neural network $\mathcal{A}$ whose depth $h=3$ and whose width of each layer is $2k^2$, where $k$ is an integer such that $k \geq d+4$. Let  $F_{\mathcal{B}}: \mathbb{R}^d \rightarrow \mathbb{R}$ be the function represented by a ReLU neural network $\mathcal{B}$ whose depth $h \leq k+2$ and whose width of each hidden layer $m_i \leq k^{3/2}$, where the parameter values of $\mathcal{B}$ are bounded by $[-b, b]$. For any constant $b>0$, there is $\epsilon >0$ such that $F_{\mathcal{A}}$ can never be approximated by $F_{\mathcal{B}}$ to $\epsilon$ $L^1$-distance. That is,
\begin{equation}\nonumber
	\int_{\mathbb{R}^d} |F_{\mathcal{A}}(x) - F_{\mathcal{B}}(x)| dx \geq \epsilon.
\end{equation}

This indicates that there exists a family of shallow ReLU neural networks which cannot be approximated by narrow networks whose depth is constrained by polynomial bounds. 

The polynomial lower bound for width efficiency is smaller than the exponential lower bound for depth efficiency~\cite{bengio2011expressive,bianchini2014complexity,montufar2014number}. That is, to approximate a deep model whose depth increases linearly, a shallow model requires at least an exponential increase in width. To approximate a shallow, wide model whose width increases linearly, a deep, narrow model requires at least a polynomial increase in depth. However, Lu~\textit{et al.}~\cite{lu2017expressive} point out that depth cannot be strictly proved to be more effective than width since a polynomial upper bound for width is still lacking. The polynomial upper bound ensures that to approximate a shallow, wide model, a deep, narrow one requires at most a polynomial increase in depth.

\subsection{Expressible Functional Space}

In addition to the studies of depth efficiency and width efficiency, the third line of works explore the classes of functions that can be expressed by deep learning models with specific frameworks and specified size. This line of works explore the expressible functional space of deep learning models, including model-specific and cross-model approaches. 

\nop{\begin{figure}
	\centering
	\subfigure[$H_{\frac{1}{2}, \frac{1}{2}}\circ \gamma_{Z(\bm{b}^1, \bm{b}^2, \bm{b}^3, \bm{b}^4)}$]{
		\includegraphics[width = 0.35\linewidth]{figures/capacity/arora2018understanding_zono1}
	}
	\hspace{.2in}
	\subfigure[$H_{\frac{1}{2}, \frac{1}{2},\frac{1}{2}}\circ \gamma_{Z(\bm{b}^1, \bm{b}^2, \bm{b}^3, \bm{b}^4)}$]{
		\includegraphics[width=0.4\linewidth]{figures/capacity/arora2018understanding_zono2}
	}
	\caption{Examples of ZONOTOPE from Arora~\textit{et al.}~\cite{arora2016understanding}.}
	\label{fig:arora2016understanding:zono}
\end{figure}}

\subsubsection{Model-specific Approaches}

Arora~\textit{et al.}~\cite{arora2016understanding} investigate the family of functions representable by deep neural networks with ReLU activation function. 
They prove that every piecewise linear function $f: \mathbb{R}^d \rightarrow \mathbb{R}$ can be represented by a ReLU neural network that consists of at most $\lceil \log_2(d+1)\rceil$ hidden layers. 
The family of piecewise linear functions is dense in the family of compactly supported continuous functions, and the family of compactly supported continuous functions is dense in Lebesgue space $L^p(\mathbb{R}^d)$~\cite{arora2016understanding,carothers2000real}. 

The Lebesgue space $L^p(\mathbb{R}^d)$ is defined as the the family of Lebesgue intergrable functions $f$ for which $\int_{\mathbb{R}^d}|f| < +\infty$~\cite{carothers2000real}. Define $L^p$ norm~\cite{carothers2000real} as $||f||_p = [\int_{\mathbb{R}^d}|f|^p]^{1/p}$. Then, the above conclusion can be extended to the $L^p(\mathbb{R}^d)$-space. That is, every function $f \in L^p(\mathbb{R}^d)$ can be approximated to arbitrary $L^p$ norm by a ReLU neural network which consists of at most $\lceil \log_2(d+1)\rceil$ hidden layers. 

G{\"u}hring~\textit{et al.}~\cite{guhring2019complexity} study the expressive capacity of deep neural networks with ReLU activation function in Sobolev space~\cite{adams2003sobolev}. Given $\Omega \subset \mathbb{R}^d$, $p \in [1, \infty]$, $n\in \mathbb{N}$, $L^p(\Omega)$ is the Lebesgue space on $\Omega$, the Sobolev space~\cite{perez2017sobolev} is defined as
\begin{equation}\nonumber
	W^{n,p}(\Omega)
=  \{f\in L^p(\Omega) : D^\alpha f \in L^p(\Omega) \mbox{ for } \forall \alpha \in \mathbb{N}^d_0, |\alpha|\leq n\},
\end{equation}
where $D^\alpha f$ is the $\alpha$-th order derivative of $f$. Sobolev norm is defined as
\begin{equation}\nonumber
	||f||_{W^{n,p}(\Omega)} = (\sum_{0\leq|\alpha|\leq n} ||D^\alpha f||^p_{L^p(\Omega)})^{1/p}.
\end{equation}

G{\"u}hring~\textit{et al.}~\cite{guhring2019complexity} analyze the effect of ReLU neural networks in approximating functions from Sobolev space, and establish upper and lower bounds on the sizes of models to approximate functions in the Sobolev space. 
Specifically, define a subset of Sobolev space $f$ as
\begin{equation}\nonumber
	\mathcal{F}_{n, d, p, B} = \{f\in W^{n,p}((0,1)^d): ||f||_{W^{n,p}((0,1)^d)} \leq B\}.
\end{equation}
The upper bound shows that, for any $d\in \mathbb{N}$, $n \in \mathbb{N}, n\geq 2$, $\varepsilon>0$, $0 \leq s \leq 1$, $1 \leq p \leq +\infty$, and $B >0$, for any function $f \in \mathcal{F}_{n, d, p, B}$, there exists a neural network $\mathcal{N}_\varepsilon$ and a choice of weights $w_f$ such that
\begin{equation}\nonumber
	||\mathcal{N}_\varepsilon(\cdot|w_f) - f(\cdot)||_{W^{s,p}((0,1)^d)} \leq \varepsilon
\end{equation}
where $\mathcal{N}_\varepsilon$ represents a ReLU neural network consisting of at most $c\log_2(\varepsilon^{-\frac{n}{n-s}})$ layers and $c\varepsilon^{-\frac{d}{n-s}}\log_2(\varepsilon^{-\frac{n}{n-s}})$ neurons, and with non-zero parameters. The value of constant $c$ depends on the value of $d, p, n, s$ and $B$.

Besides, the lower bound~\cite{guhring2019complexity} shows that, for any  $d\in \mathbb{N}$, $n \in \mathbb{N}, n\geq 2$, $\varepsilon>0$, $B >0$, $k \in \{0, 1\}$, with $p = +\infty$, for any function $f \in \mathcal{F}_{n, d, p, B}$, there exists a ReLU neural network $\mathcal{N}_\varepsilon$ such that
\begin{equation}\nonumber
	||\mathcal{N}_\varepsilon(\cdot|w_f) - f(\cdot)||_{W^{k,\infty}((0,1)^d)} \leq \varepsilon
\end{equation}
where $\mathcal{N}_\varepsilon$ has at least $c'\varepsilon^{-\frac{d}{2(n-1)}}$ non-zero weights.

Kileel~\textit{et al.}~\cite{kileel2019expressive} explore the functional space of deep neural networks with polynomial activation functions. A polynomial activation function $\rho_r(z)$ raises $z$ to the power of $r$. 
They suggest that, with polynomial activation functions, the study of model complexity can be benefitted from the application of powerful mathematical machinery of algebraic geometry. In addition, polynomials can approximate any continuous activation function, thus help to explore other deep learning models. 

Given a deep polynomial neural network $\mathcal{N}: \mathbb{R}^{d} \rightarrow \mathbb{R}^{c}$ with depth $h$ and polynomial degree $r$, let $m = \{m_0, m_1,\ldots, m_h\}$ represent the architecture of $\mathcal{N}$ with the width $m_i$ of layer $i$ and $m_0=d$, $m_h=c$. Let $\mathcal{F}_{m,r}$ be the functional space of $\mathcal{N}$. 
Kileel~\textit{et al.}~\cite{kileel2019expressive} define the functional variety $\mathcal{V}_{m,r}$ as $\mathcal{V}_{m,r} = \overline{\mathcal{F}_{m,r}}$, which is the Zariski closure of the functional space $\mathcal{F}_{m,r}$. ``The Zariski closure of a set $X$ is the smallest set containing $X$ that can be described by polynomial equations.''~\cite{kileel2019expressive} The functional variaty $\mathcal{V}_{m,r}$ can be much larger than the functional space $\mathcal{F}_{m,r}$. 
Kileel~\textit{et al.}~\cite{kileel2019expressive} propose to use the dimensionality of functional variety, denoted by $\dim \mathcal{V}_{m,r}$, as the representation of the expressive capacity of deep polynomial neural networks, written as
\begin{equation}\nonumber
	MEC(\mathcal{N}_{m, r}) = \dim \mathcal{V}_{m,r}.
\end{equation}

To measure $\dim \mathcal{V}_{m,r}$, Kileel~\textit{et al.}~\cite{kileel2019expressive} build connections between deep polynomial networks and tensor decompositions. Specifically, polynomial networks with $h=2$ are connected to CP tensor decompositions~\cite{landsberg2012tensors}, and deep polynomial networks are connected to an iterated tensor decomposition~\cite{lundqvist2019generic}. 
Based on decompositions, they prove that, for any fixed $m$, there exists $\tilde{r} = r(m)$ such that for any $r > \tilde{r}$,  $\dim \mathcal{V}_{m,r}$  is bounded by
\begin{equation}\nonumber
 \dim \mathcal{V}_{m,r}
\leq  \min \left(m_h+\sum_{i=1}^{h}(m_{i-1}-1)m_i, m_h\binom{m_0+r^{h-1}+1}{r^{h-1}} \right).
\end{equation}

Besides, Kileel~\textit{et al.}~\cite{kileel2019expressive} prove a bottleneck property of deep polynomial networks. That is, a too narrow layer is a bottleneck and may ``chock'' the polynomial network such that the network can never fill the ambient space. The ambient space $Sym_r(\mathbb{R}^d)$ is the space of homogeneous polynomials of degree $r$ in $d$ variables. A polynomial network filling the ambient space satisfies $\mathcal{F}_{m, r}=Sym_{r^{h-1}}(\mathbb{R}^d)^c$. They show that, network architectures filling the ambient space can be helpful for optimization and training. 
 
\subsubsection{Cross-Model Approaches}

In addition to model-specific approaches, expressible functional space can be investigated in a cross-model manner.
Specifically, Khrulkov~\textit{et al.}~\cite{khrulkov2017expressive} study the expressive capacity of recurrent neural networks (RNNs). They investigate the connections between network architectures and tensor decompositions, then make comparison between the expressive capacity of RNNs, CNNs, and shallow FCNNs. 

\begin{figure*}[t]
	\centering
	\subfigure[TT-decomposition]{
		\label{fig:ttdecomposition}
		\includegraphics[width = 0.26\linewidth]{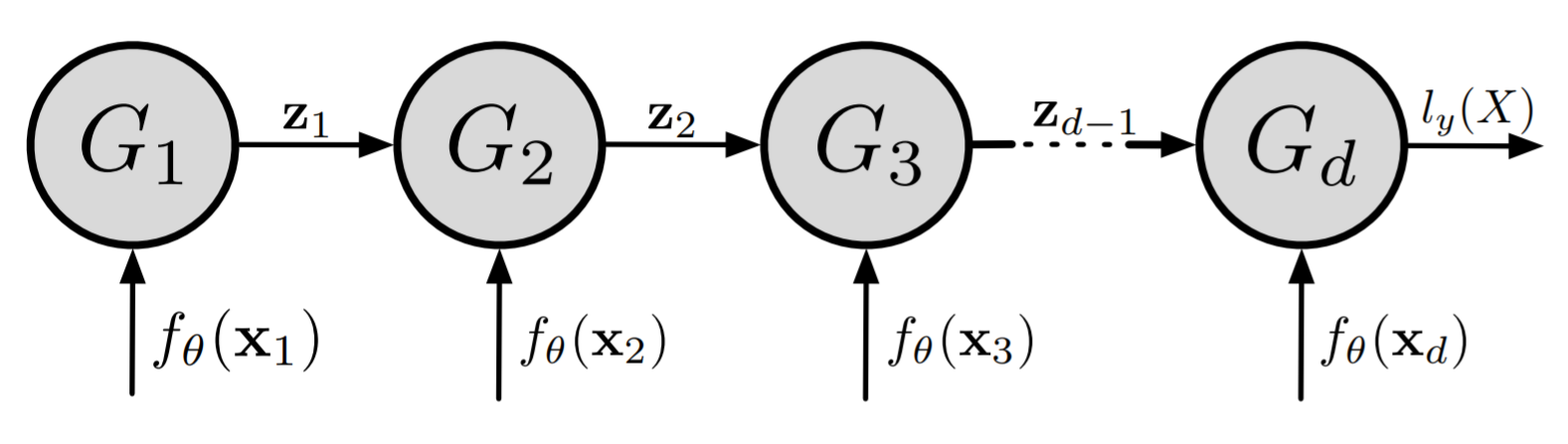}
	}
	\hspace{.1in}
	\subfigure[CP-decomposition]{
		\label{fig:cpdecomposition}
		\includegraphics[width=0.3\linewidth]{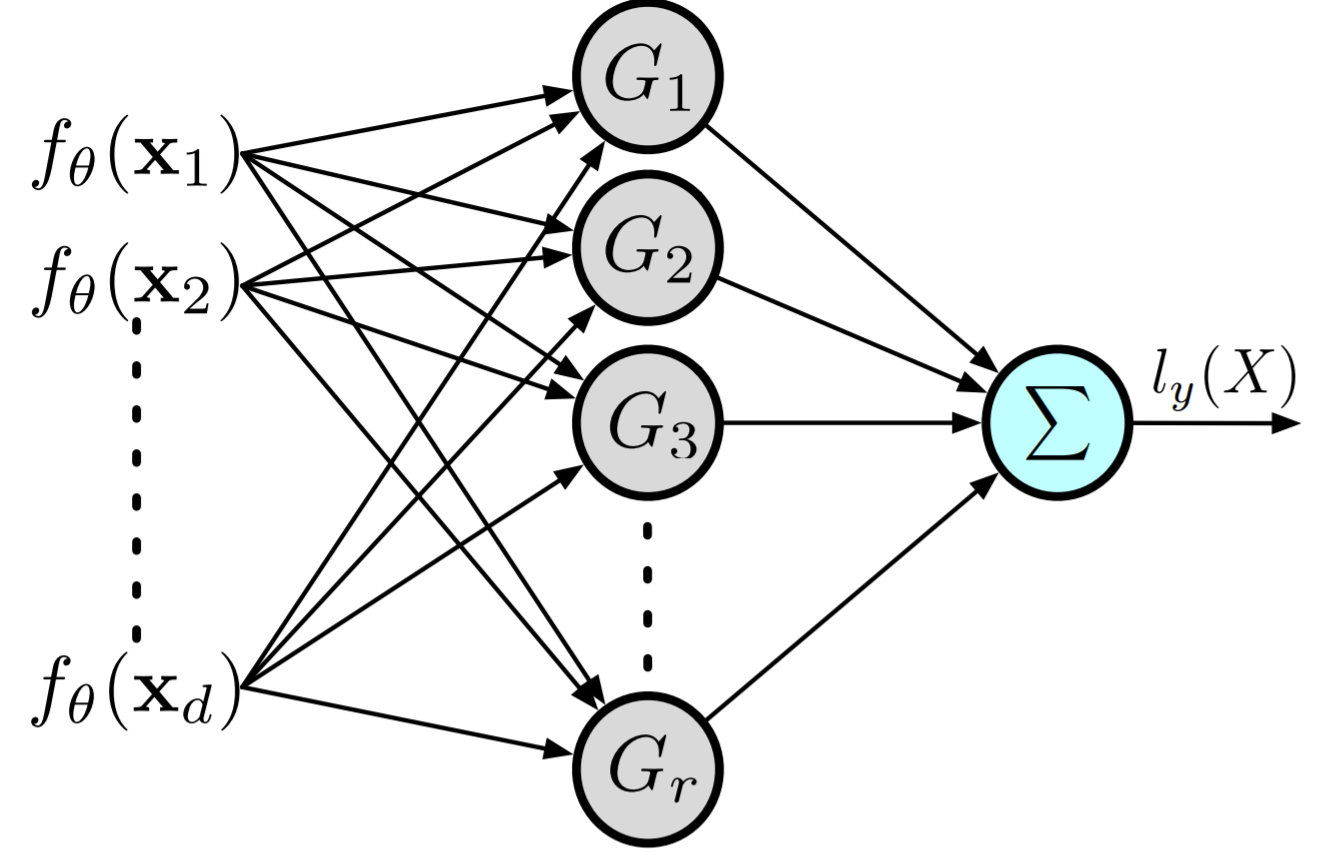}
	}
	\hspace{.1in}
	\subfigure[HT-decomposition]{
		\label{fig:htdecomposition}
		\includegraphics[width=0.3\linewidth]{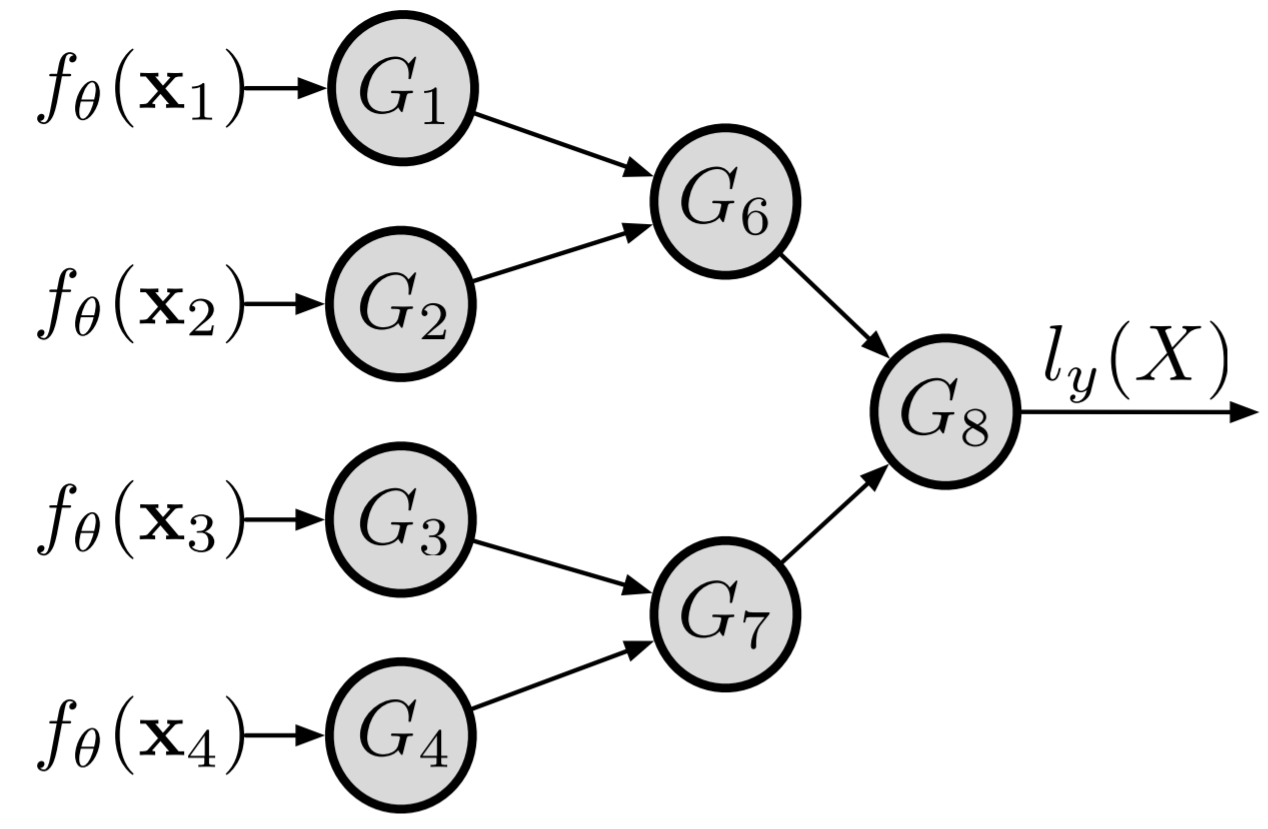}
	}
	\caption{Examples of networks corresponding to various tensor decompositions~\cite{khrulkov2017expressive}, from the left are TT decomposition, CP-decomposition, HT-decomposition.}
\end{figure*}
Let $\mathcal{X} \in \mathbb{R}^{n_1\times n_2 \times \ldots \times n_d}$ be a $d$-dimensional tensor. The Tensor Train (TT) decomposition~\cite{oseledets2011tensor} of a tensor $\mathcal{X}$ is computed by 
\begin{equation}\nonumber
	\mathcal{X}^{i_1i_2\ldots i_d} = \sum_{\alpha_1=1}^{r_1} \nop{\sum_{\alpha_2=1}^{r_2}} \ldots \sum_{\alpha_{d-1}=1}^{r_{d-1}} G_1^{i_1\alpha_1} G_2^{\alpha_1i_2\alpha_2} \ldots G_d^{\alpha_{d-1}i_d}
\end{equation}
where the tensors $G_k \in \mathbb{R}^{r_{k-1}\times n_k\times r_k}$ are called TT-cores. Khrulkov~\textit{et al.}~\cite{khrulkov2017expressive} introduce bilinear units to represent TT-cores. Given $x\in \mathbb{R}^m, y\in \mathbb{R}^n$ and $G\in \mathbb{R}^{m\times n\times k}$, a bilinear unit performs a map $G: \mathbb{R}^m \times \mathbb{R}^n \rightarrow \mathbb{R}^k$, written as $G(x,y) = z$. Based on the bilinear units, they show that a recurrent neural network realizes the TT-decomposition of the weight tensor~(Fig.~\ref{fig:ttdecomposition}).
Similarly, Khrulkov~\textit{et al.}~\cite{khrulkov2017expressive} show that the Canonical (CP) decomposition~\cite{carroll1970analysis}, with the form of
\begin{equation}\nonumber
	\mathcal{X}^{i_1i_2\ldots i_d} = \sum_{\alpha = 1}^{r} v_{1,\alpha}^{i_1} v_{2, \alpha}^{i_2} \ldots v_{d,\alpha}^{i_d}, 
\end{equation}
corresponds to a one-hidden-layer FCNN~(Fig.~\ref{fig:cpdecomposition}). Each unit of the network is denoted by $G_\alpha = v_{1,\alpha}^{i_1} v_{2, \alpha}^{i_2} \ldots v_{d,\alpha}^{i_d}$, where $v_{i,\alpha} \in \mathbb{R}^{n_i}$. The Hierarchical Tucker~(HT) decomposition~\cite{cohen2016convolutional} corresponds to an CNN structure~(Fig.~\ref{fig:htdecomposition}). 

\begin{table}[t]
	\begin{center}
		\caption{Comparison of the expressive power of various network architectures given by Khrulkov~\textit{et al.}~\cite{khrulkov2017expressive}. Each column represents a specific network with width $r$, rows show the upper bounds on the width of other types of networks to achieve equivalent expressive capacity.} 
		\label{tab:tensorcomparison}
		\begin{tabular}{cccc}
			\hline
			& TT-Network & HT-Network & CP-Network \\
			\hline
			TT-Network & $r$ & $r^{\log_2(d)/2}$ & $r$ \\
			HT-Network & $r^2$ & $r$ & $r$ \\
			CP-Network & $\geq r^{d/2}$ & $\geq r^{d/2}$ & $r$ \\
			\hline
		\end{tabular}
	\end{center}
\end{table}

Khrulkov~\textit{et al.}~\cite{khrulkov2017expressive} propose the rank of tensor decomposition as a measure of neural network complexity, since the rank of decompositions corresponds to the width of networks. Based on the correspondence relationships between neural networks and tensor decompositions, they compare the model complexity of RNNs, CNNs, and shallow FCNNs.
The main conclusions are summarized in Table~\ref{tab:tensorcomparison}. In particular, given a random $d$-dimensional tensor whose TT-decomposition is with rank $r$ and mode size $n$, this tensor has exponentially large ranks of CP-decomposition and HT-decomposition. That tells, to approximate a recurrent neural network, a shallow FCNN or CNN requires an exponentially larger width.

\subsection{VC Dimension and Rademacher Complexity}
\label{sec:capacitydiscussion}

\nop{In this subsection, we further discuss two specific problems of the expressive capacity of deep learning models. First, we discuss the VC dimension and Rademacher complexity of deep learning models. Second, we discuss the explanations of depth efficiency. }

The VC dimension and Rademacher complexity are widely used to analyze the expressive capacity and generalization of classical parametric machine learning models~\cite{bartlett2002model,bartlett2002rademacher,cherkassky1999model,mohri2018foundations,tan2004support}. A series of works investigate the VC dimension and Rademacher complexity of deep learning models. 

The VC dimension is an expressive capacity measure that reflects the largest number of samples that can be shattered by the hypothesis space~\cite{mohri2018foundations}. A higher VC dimension means the model can shatter a larger number of samples and thus the model has a higher expressive capacity. Maass~\cite{maass1994neural} studies the VC dimension of feedforward neural networks with linear threshold gates. The linear threshold gate means that each neuron is composed of a weighted sum function followed by a Heaviside activation function (Eq.~(\ref{eq:heaviside})). Let $W$ be the number of parameters in the network and $L$ be the depth of the network. Maass~\cite{maass1994neural} proves that for $L \geq 3$, the VC dimension of such networks is $\Theta(W\log W)$. 

Bartlett~\textit{et al.}~\cite{bartlett1998almost} investigate the VC dimension of feedforward neural networks with piecewise polynomial activation functions. A piecewise polynimial activation function with $p$ pieces has the form $\sigma(z) = \phi_i(z)$, where $z\in[t_{i-1}, t_i)$, $i \in {1,\ldots,p+1}$, and $t_{i-1}< t_{i}$. Each $\phi_i$ is a polynomial function of degree no more than $r$. Let $W$ be the number of parameters in the network and $L$ be the depth of the network. Bartlett~\textit{et al.}~\cite{bartlett1998almost} prove that the upper bound for the VC dimension of such networks is $O(WL^2+WL\log WL)$ and the lower bound for the VC dimension is $\Omega(WL)$. Later, Bartlett~\textit{et al.}~\cite{bartlett2019nearly} improve this lower bound to $\Omega(WL\log(W/L))$.

Bartlett~\textit{et al.}~\cite{bartlett2019nearly} study the VC dimension for deep neural networks with piecewise linear activation functions (e.g., ReLU).  Given a deep neural network with $L$ layers and $W$ parameters, they prove that the lower bound for VC dimension of such networks is $\Omega(WL\log (W/L))$ and the upper bound for VC dimension is $O(WL\log W)$.

Rademacher complexity captures the capacity of a hypothesis space to fit random labels as a measure of the expressive capacity~\cite{mohri2018foundations}. A higher Rademacher complexity means that the model can fit a larger number of random labels and thus the model has a higher expressive capacity.
Bartlett~\textit{et al.}~\cite{bartlett2017spectrally} investigate the Rademacher complexity of deep neural networks with ReLU activation function. Given a deep ReLU neural network with $L$ layers, let $A_i$ be the parameter matrix of layer $i$, and $X\in \mathbb{R}^{n\times d}$ the data matrix, where $n$ is the number of instances and $d$ is the input dimension. Bartlett~\textit{et al.}~\cite{bartlett2017spectrally} prove that the lower bound for the Rademacher complexity of such networks is $\Omega(||X||_F\prod_i ||A_i||_\sigma)$, where $||\cdot||_\sigma$ is the Spectral norm, and $||\cdot||_F$ is the Frobenius norm. 
Neyshabur~\textit{et al.}~\cite{neyshabur2018role} prove a tighter lower bound for two-layer ReLU neural networks. Suppose $||A_1||_\sigma \leq s_1$, $||A_2||_\sigma \leq s_2$, and $s_1s_2$ is the Lipschitz bound of the function represented by the network. They prove that the Rademacher complexity is lower bounded by $\Omega(s_1s_2\sqrt{m}||X||_F/n)$, where $m$ is the width of the hidden layer. This lower bound improves the bound~\cite{bartlett2017spectrally} by a factor of $\sqrt{m}$.

Yin~\textit{et al.}~\cite{yin2019rademacher} study the Rademacher complexity of adversarial trained neural networks. Given a feedforward neural network with ReLU activation function, denoted by $f$, let $L$ be the depth of the network, and $A_i$ the parameter matrix of layer $i$. The function family represented by $f$ with adversarial loss function can be written as
\begin{equation}\nonumber
	\mathcal{F} = \{f_A: \min_{x'\in\mathbb{B}(\epsilon)} yf_A(x), \prod _{i=1}^L ||A_i||_\sigma \leq r\}
\end{equation}
where $\mathbb{B}(\epsilon) = \{x'\in \mathbb{R}^d: ||x' - x||\leq \epsilon\}$ is the set of samples perturbed around $x$ with $l_\infty$ distance $\leq \epsilon$. 
Yin~\textit{et al.}~\cite{yin2019rademacher} prove that the lower bound for the Rademacher complexity of $\mathcal{F}$ is $\Omega(||X||_F/n + \epsilon\sqrt{d/n})$. This lower bound  exhibits an explicit dependence on the input dimension $d$.


Some studies~\cite{neyshabur2017implicit,zhang2016understanding} suggest that deep learning models are often over-parameterized in practice and have significantly more parameters than samples. In this case, the VC dimension and Rademacher complexity of deep learning models are always too high, so the practical guidance they can provide is weak.





\section{Effective Complexity of Deep Learning Models}
\label{sec:effective}
Effective complexity of deep learning models is also known as practical complexity, practical expressivity, and usable capacity~\cite{hanin2019complexity,novak2018sensitivity}. It reflects the complexity of the functions represented by deep models with specific parameterizations~\cite{raghu2017expressive}. 
	
Effective complexity of deep learning models has been mainly explored from two different aspects.
\begin{itemize}
	\item \textbf{General measures of effective complexity}\nop{ the studies from this aspect} design quantitative measures for effective complexity of deep learning models.
	\item Investigations into the \textbf{high-capacity low-reality phenomenon} find that effective complexity of deep learning models may be far lower than their expressive capacity. 
\end{itemize}
In this section, we review these two groups of studies. 

\subsection{General Measures of Effective Complexity}

Compared to expressive capacity, the study of effective model complexity has stronger requirements for sensitive and precise complexity measures. This is because the effective complexity cannot be directly derived from the model structure alone~\cite{hu2020lann}. Different parameter values of the same model structure may lead to different effective complexity. An effective complexity measure is expected to be sensitive to different parameter values used in models with the same structure.

A series of works devote to proposing feasible measures for effective complexity of deep learning models. A major group of the complexity measures depends on the linear region splitting of piecewise linear neural networks in the input space.  We start with those methods and then discuss the others.

\subsubsection{Piecewise Linear Property}

\begin{figure}[t]
	\centering
	\includegraphics[width = 0.6\linewidth]{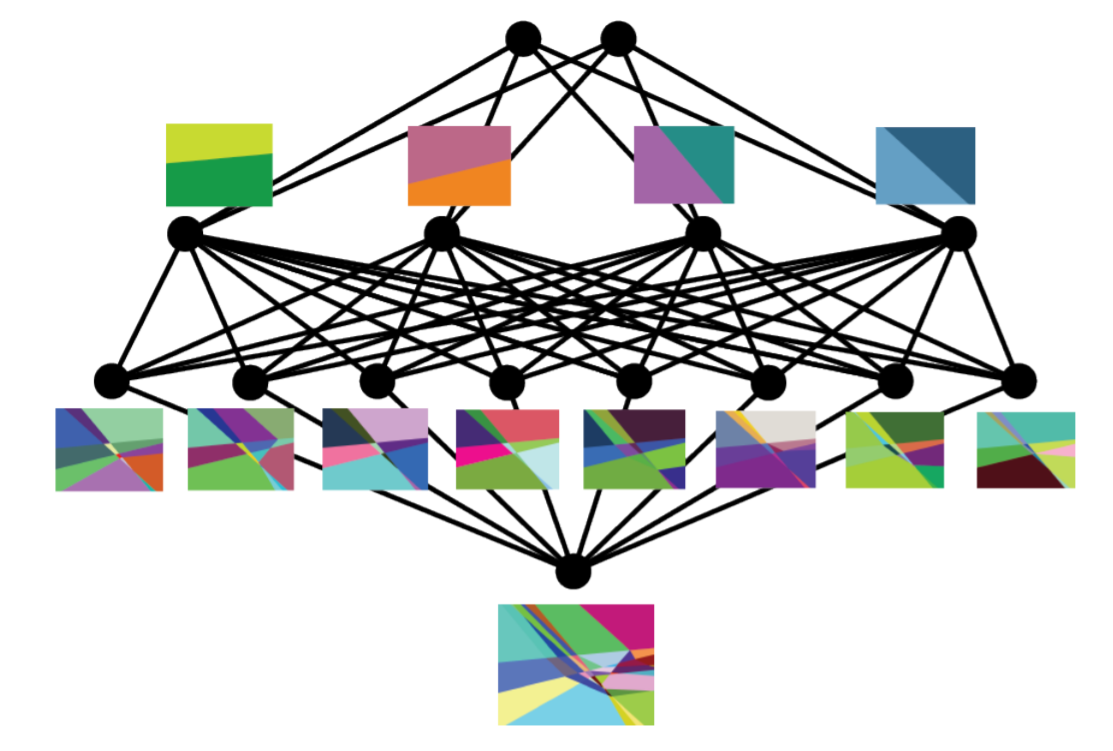}
	\caption{A 2-hidden-layer ReLU network divides the 2-dimensional input space into a number of linear regions, studied by Hanin and Rolnick~\cite{hanin2019complexity}. Upon each hidden neuron shows the linear regions divided by the current neuron.}
	\label{fig:hanin2019complexity:nn}
\end{figure}

It is well known that a neural network with piecewise linear activation functions generates a finite number of linear regions in the input space~\cite{hu2020lann,montufar2014number,raghu2017expressive}. This property is called the \emph{piecewise linear property} and is demonstrated in Fig.~\ref{fig:hanin2019complexity:nn}. The number of linear regions as well as the density of such regions can usually reflect the effective complexity. Therefore, a series of studies on effective complexity starts from piecewise linear activation functions (e.g., ReLU, Maxout) or are based on the piecewise linear property.

Raghu~\textit{et~al.}~\cite{raghu2017expressive} propose two interrelated effective complexity measures for deep neural networks with piecewise linear activation functions, such as ReLU and hard Tanh~\cite{nwankpa2018activation}. 
Specifically, they define trajectory $x(t)$ between two input points $x_1, x_2 \in \mathbb{R}^d$, which is a curve parametrized by a scalar $t \in [0,1]$, with $x(0)=x_1$ and $x(1)=x_2$. The first effective complexity measure they propose is the number of linear regions in the input space when sweeping an input point through a trajectory path $x(t)$, that is, the effective complexity $EMC(\mathcal{N})$ of model $\mathcal{N}$ is
\begin{equation}\nonumber
	EMC(\mathcal{N}) = \mathcal{T}(\mathcal{N}(x(t); W))
\end{equation}
where $\mathcal{T}$ is the number of linear regions passing through the trajectory $x(t)$ and $W$ is a specific set of model parameters.
The second effective complexity measure they propose  is the trajectory length $l(x(t))$ defined as
\begin{equation}\nonumber
	l(x(t)) = \int_t ||\frac{dx(t)}{dt}||dt
\end{equation}
which is the standard \emph{arc length} of the trajectory $x(t)$. They prove the proportional relationship between these two complexity measures. 
To obtain the estimated value of effective complexity, Raghu~\textit{et~al.}~\cite{raghu2017expressive} bound the expected value of trajectory length of any layer in a deep ReLU neural network.
Specifically, given a deep ReLU neural network whose weights are initialized by $N(0, \sigma^2_w/m)$ and the biases are initialized by $N(0, \sigma_b^2)$. Let $z^{(i)}(x(t))$ denote the new trajectory obtained after the transformation of the first $i$ hidden layers of the trajectory $x(t)$. The expected value of its trajectory length can be bounded by
\begin{equation}\nonumber
	\mathbb{E}[l(z^{(i)}(x(t)))] \geq O(\frac{\sigma_w \sqrt{m}}{\sqrt{m+1}})^i l(x(t))
\end{equation}
where $m$ denotes the hidden layer width. Similarly, for a hard Tanh neural network under the same initialization, the trajectory length is bounded by
\begin{equation}\nonumber
	\mathbb{E}[l(z^{(i)}(x(t)))] \geq O\left( \frac{\sigma_w \sqrt{m}}{\sqrt{\sigma_w^2 + \sigma_b^2 + m\sqrt{\sigma_w^2 + \sigma_b^2}}}
	\right)^i  l(x(t)).
\end{equation}

\begin{figure}[t]
	\centering
	\includegraphics[width = 0.8\linewidth]{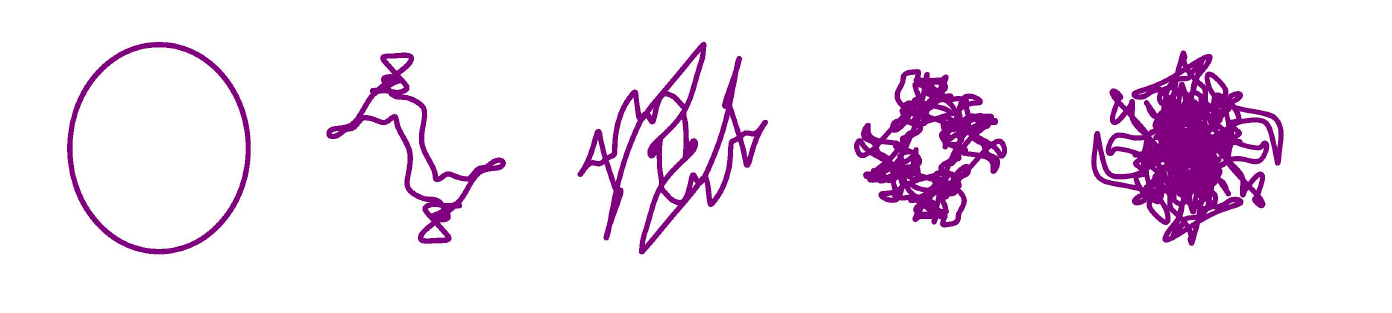}
	\caption{A circular trajectory (the left most) is fed into a Tanh neural network, then the following images show the trajectory after the transformation of each hidden layer in turn~\cite{raghu2017expressive}. It shows the increase of the length of the trajectory after the layer transformation. }
	\label{fig:raghu2017expressive:traje}
\end{figure}
\nop{\begin{figure}
	\centering
	\subfigure[Without Batch Norm]{
		\includegraphics[width = 0.38\linewidth]{figures/effective/raghu2017expressive_grow1}
	}
	~~
	\subfigure[With Batch Norm]{
		\includegraphics[width = 0.45\linewidth]{figures/effective/raghu2017expressive_grow2}
	}
	\caption{Growth of the length of a circular trajectory between two datapoints with or without Batch Norm layers for a convolutional neural network on CIFAR-10. }
	\label{fig:raghu2017expressive:grow}
\end{figure}}

Using these two complexity measures, Raghu~\textit{et~al.}~\cite{raghu2017expressive} explore the performance of deep neural networks and report some findings. First, the effective complexity grows exponentially with respect to the depth of the model and polynomially with respect to the width. Fig.~\ref{fig:raghu2017expressive:traje} is an example showing the evolution of a trajectory after each hidden layer of a deep network. Second, how the parameters are initialized affects the effective complexity. Third, injecting perturbations to a layer leads to exponentially larger perturbations at the remaining layers. Last, regularization approaches, such as Batch Normalization~\cite{ioffe2015batch}, help to reduce trajectory length. This explains why Batch Normalization helps model stability and generalization.

Novak~\textit{et~al.}~\cite{novak2018sensitivity} empirically investigate the relationship between model complexity and generalization of neural networks with piecewise linear activation functions. They propose to use model sensitivity to measure effective complexity. Model sensitivity, also known as robustness, reflects the capacity of a model to distinguish between different inputs at small distances. Novak~\textit{et~al.}~\cite{novak2018sensitivity} introduce two sensitivity metrics, input-output Jacobian norm and trajectory length~\cite{raghu2017expressive}. 

First, based on the piecewise linear property, 
Novak~\textit{et~al.}~\cite{novak2018sensitivity} propose the Jacobian norm to measure the local sensitivity under the assumption that the input is perturbed within the same linear region. This Jacobian norm can further represent the effective model complexity.
That is, 
\begin{equation}\nonumber
	EMC(\mathcal{N}) = \mathbb{E}_{{x}\sim D} [|| J({x})||_F]
\end{equation}
where $J(x) = \partial f_\mathcal{N} (x)/\partial x^T$ is the Jacobian of class probabilities, $f_\mathcal{N}$ is the function represented by the network $\mathcal{N}$, $D$ is the data distribution, and $||\cdot||_F$ is the Frobenius norm. 

Second, they propose a trajectory length metric similar to what developed by Raghu~\textit{et~al.}~\cite{raghu2017expressive} as a sensitivity measure when the input is perturbed to other linear regions, further as a measure of effective complexity, written as
\begin{equation}\nonumber
	EMC(\mathcal{N}) = \mathbb{E}_{{x}\sim D}[t({x})]
\end{equation} 
where $t(x)$ is a trajectory length defined by the number of linear regions passing through a trajectory $\tau (x)$, that is,
\begin{equation}\nonumber
	\begin{split}
		t(x) & = \sum_{i=0}^{k-1} ||c(z_i) - c(z_{(i+1)\%k})||_1 \\
			  & \approx \oint_{z \in \tau(x)} ||\frac{\partial c(z)}{\partial (dz)}||_1 dz
	\end{split}
\end{equation}
where $z_0, \ldots, z_{k-1}$ are $k$ equidistant points on the trajectory $\tau (x)$, $c(z)$ is the status encoding of all hidden neurons at point $z$. 

Using these two complexity measures, Novak~\textit{et~al.}~\cite{novak2018sensitivity} study the correlation between complexity and generalization. They demonstrate that neural networks have strong robustness in the vicinity of the training data manifold where deep models have good generalization capability. They also show that the factors associated with poor generalization (e.g., full-batch training, random labels) correspond to weaker robustness, and the factors associated with good generalization (e.g., data augmentation, ReLU) correspond to stronger robustness.

To develop a measure of effective complexity for general smooth activation functions, Hu~\textit{et~al.}~\cite{hu2020lann} propose an effective complexity measure for deep neural networks with curve activation functions (e.g., Sigmoid~\cite{kilian1993power}, Tanh~\cite{kalman1992tanh}). Motivated by the piecewise linear property, they suggest that, using a piecewise linear function with a minimal number of linear regions to approximate a given network, the number of linear regions of the approximation function can be a measure of effective complexity of the given network. 
They learn a piecewise linear approximation of a deep neural network with curve activation functions, which is called the Linear Approximation Neural Network (LANN for short). The LANN is constructed by learning a piecewise linear approximation function for the curve activation function on each hidden neuron. 
Specifically, 
Hu~\textit{et~al.}~\cite{hu2020lann} define the approximation error of a LANN $g$ to the target network $f$ as $\mathcal{E}(g;f) = \mathbb{E}[|g(x) - f(x)|]$.
They analyze how the approximation error at a certain layer is propagated through the remaining hidden layers, and obtain the influence of the approximation error of each hidden neuron on $\mathcal{E}(g;f)$. That is, 
\begin{equation}\nonumber
	\small
	\label{eq:hu2020lann}
	\mathcal{E}(g; f) = \sum_{i,j} \frac{1}{c}\sum(\ |V_o|\prod_{q=L}^{i+1} \mathbb{E}[|J_q|]\ )_{*, j} (\mathbb{E}[e_{i,j}] + \mathbb{E}[|\hat{\epsilon}_{i,j}|])
\end{equation}
where $J_q$ is the Jacobian matrix of layer $q$ of the network $f$, $e_{i,j}$ is the approximation error of $g$ on a specific neuron $\{i,j\}$, $V_o$ is the weight matrix of the output layer, and $\hat{\epsilon}_{i,j}$ is the negligible estimation error on neuron $\{i,j\}$. Given an approximation degree $\lambda$, the LANN with the smallest number of linear regions is constructed to meet the requirement $\mathcal{E}(g;f) \leq \lambda$.  
The approximated number of linear regions of the LANN is used as the effective complexity measure, that is,
\begin{equation}\nonumber
	EMC(f)_\lambda = d \sum_{i=1}^{L} \log (\sum_{j = 1}^{m_i} k_{i,j} - m_i + 1)
\end{equation}
where $k_{i,j}$ is the number of linear pieces of the approximation function on neuron $\{i,j\}$, $m_i$ is the width of layer $i$, and $L$ is the depth of $f$.

Using the complexity measure, Hu~\textit{et~al.}~\cite{hu2020lann} investigate the trend of model complexity in the training process. They show that the effective complexity increases with respect to the number of training iterations. They also demonstrate that the occurrence of overfitting is positively correlated with the increase of effective complexity, while regularization methods (e.g., $L^1, L^2$ regularizations) suppress the increase of model complexity (see Fig.~\ref{fig:hu2020lann:moon}).

\begin{figure}[t]
	\centering
	\includegraphics[width = 0.6\linewidth]{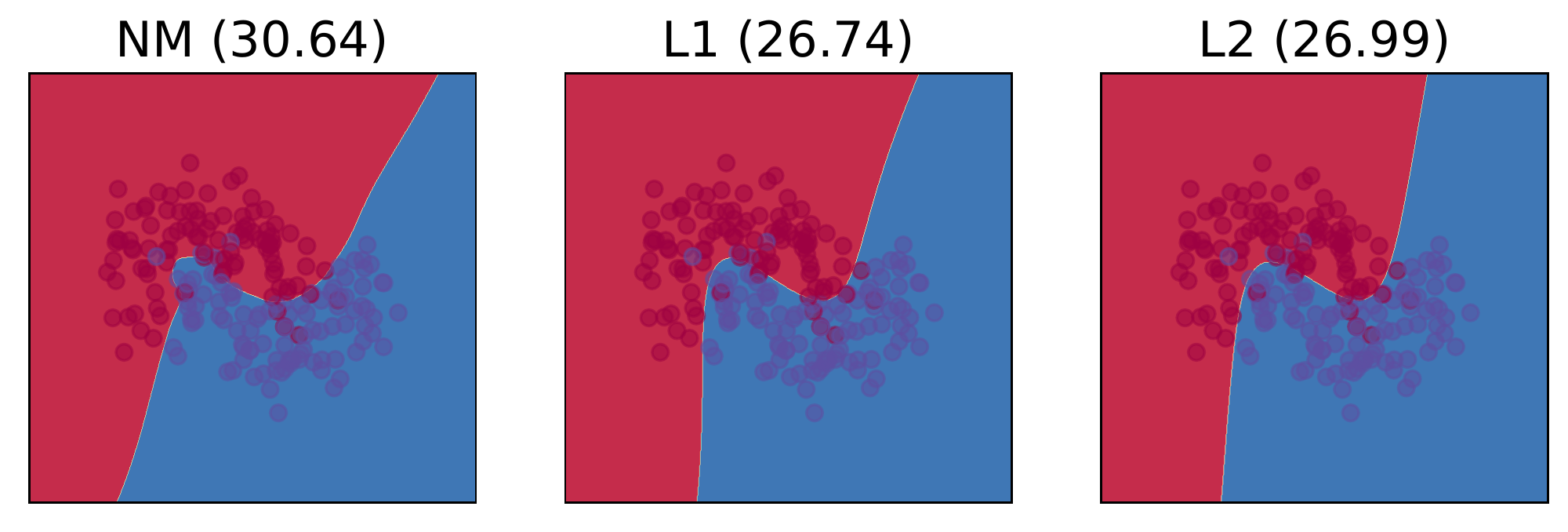}
	\caption{Influence of regularization approaches on effective complexity~\cite{hu2020lann}. This figure shows that the decision boundaries of models trained on the Moon dataset. NM, L1, L2 are short for normal train, train with $L^1$, and $L^2$ regularization, respectively. In brackets are the value of effective complexity measure. }
	\label{fig:hu2020lann:moon}
\end{figure}

The piecewise linear property may provide novel opportunities for capturing model complexity in deep learning. In addition to the above studies on effective complexity, piecewise linear neural networks or the piecewise linear property can also help the exploration of expressive capacity (see Section 2, \cite{arora2009computational,guhring2019complexity,lu2017expressive,montufar2014number,serra2018bounding}).
Piecewise linear activation functions, especially ReLU, are popular and effective activation functions in many tasks and applications~\cite{lecun2015deep,montufar2014number}. The local linear characteristic and a finite number of regional divisions facilitate quantifying and analyzing neural network model complexity with piecewise linear activation functions.


\subsubsection{Other Measure Metrics}

There are other effective complexity measures based on ideas other than the piecewise linear property. 

Nakkiran~\textit{et~al.}~\cite{nakkiran2019deep} introduce an effective complexity measure by investigating the double descent phenomenon. The double descent phenomenon of deep neural networks is that, as the model size, training epochs, or size of training data increases, the test performance often first decreases and then  increases. 
They suggest that to help capture the double descent effect in training, a complexity measure should be sensitive to training processes, data distribution, and model architectures. 
They propose such an effective complexity measure, that is, the maximum number of samples on which the model can be trained to close to zero training error, written as
\begin{equation}\nonumber
	EMC_{D, \epsilon}(\mathcal{T})= \max\ \{n | \mathbb{E}_{S\sim D}[Error_S(\mathcal{T}(S))] \leq \epsilon\}
\end{equation}
where $D$ is the data distribution, $\epsilon$ is the threshold of training error, $\mathcal{T}$ is the training procedure, $Error_S$ is the mean error of the model on the training set $S$, and $n$ the size of training set $S$. 
Based on the effective complexity measure $EMC_{D, \epsilon}(\mathcal{T})$, they investigate the evolution of the training process and show that, if $EMC_{D,\epsilon}(\mathcal{T})$ is sufficiently smaller or sufficiently larger than $n$, the size of the training set, then any perturbation of $\mathcal{T}$ that increases its effective complexity $EMC_{D, \epsilon}(\mathcal{T})$ helps to improve the test performance. However, if $EMC_{D,\epsilon}(\mathcal{T})\approx n$, then any perturbation of $\mathcal{T}$ that increases its effective complexity  $EMC_{D, \epsilon}(\mathcal{T})$ hurts the test performance.

To approach the generalization problem of deep learning models, Liang~\textit{et~al.}~\cite{liang2017fisher} introduce a new notion of model complexity measure, the Fisher-Rao norm. Their study focuses on deep fully connected neural networks whose activation function $\sigma(\cdot)$ satisfies $\sigma(z) = \sigma'(z)z$. The model complexity represented by Fisher-Rao norm is given by
\begin{equation}\nonumber
EMC(\mathcal{N}) = ||\theta||^2_{fr} = \langle\theta, I(\theta)\theta\rangle
\end{equation}
where $\theta$ is the set of parameters (i.e., weights, bias) of neural network $\mathcal{N}$, $\langle\cdot, \cdot\rangle$ is the inner product, and $I(\theta)$ is the Fisher information matrix, written as
\begin{equation}\nonumber
I(\theta) = \mathbb{E}[\nabla_\theta \ell(\mathcal{N} (X), Y) \otimes \nabla_\theta \ell(\mathcal{N} (X), Y)]
\end{equation}
where $\ell(\cdot, \cdot)$ is the loss function and $\otimes$ is the tensor product. 
Liang~\textit{et~al.}~\cite{liang2017fisher} introduce the geometric invariance property that a complexity measure should satisfy in order to study generalization capability. 
The invariance property essentially says that, since many different continuous operations may lead to exactly the same prediction, thus the generalization should only depend on the equivalence classes obtained by these continuous transformations. Specific parameterization should not affect the generalization. The complexity measure used to investigate generalization should satisfy the invariance property.  They demonstrate that this Fisher-Rao norm satisfies the invariance property. Let $\theta_1, \theta_2$ denote two parameter settings of a model $f$. If $f_{\theta_1} = f_{\theta_2}$, then their Fisher-Rao norms are equal, that is, $||\theta_1||_{fr} = ||\theta_2||_{fr}$. In particular, the Fisher-Rao norm remains invariant during the node-wise rescaling of a model. 

Effective complexity can be measured by the size of training samples that a model achieves zero training error~\cite{nakkiran2019deep}, or by the Fisher-Rao metric~\cite{liang2017fisher}. More effective complexity measures along the line may be developed. 


\subsection{High-Capacity Low-Reality Phenomenon}

Several studies explore the gap between the effective complexity and the expressive capacity of deep learning models. 

Ba and Caruana~\cite{ba2014deep} show that shallow fully connected neural networks can learn complex functions previously learned by deep neural networks, sometimes even only requiring the same number of parameters as the deep networks. 
Specifically, given a well-trained deep model, they propose to train a shallow model based on the outputs of the deep model, to mimic the deep model. They show that, the shallow mimic model can achieve an accuracy as high as the deep model. However, the shallow model cannot be trained directly on the original labeled training data to achieve the same accuracy.  This is also well recognized as knowledge distillation~\cite{hinton2015distilling}.

Based on this phenomenon, Ba and Caruana~\cite{ba2014deep} conjecture that the strength of deep learning may arise in part from a good match between deep architectures and current training algorithms. That is, compared with shallow architectures, deep architectures may be easier to train by the current optimization techniques. Moreover, they propose that when it is able to mimic the function learned by a deep complex model using a shallow model, the function learned by the deep model is not really too complicated to learn. This study suggests that there may be a big gap between the practical effective complexity of a deep learning model and the theoretical bound of its expressive capacity. We call this the \emph{high-capacity low-reality phenomenon}. 

Hanin and Rolnick~\cite{hanin2019complexity} investigate the high-capacity low-reality phenomenon in fully connected neural networks with piecewise linear activation functions, especially ReLU. They propose two effective complexity measures using the number of linear regions in the input space and the volume of boundaries between these linear regions. 

First, they investigate ReLU neural networks whose dimensionality of input and that of output are both equal to 1, and use the number of linear regions as the effective complexity measure. They show that the average number of linear regions grows linearly with respect to the total number of neurons, far below the exponential upper bound. 
Specifically, given a ReLU neural network $\mathcal{N}: \mathbb{R}\rightarrow \mathbb{R}$ whose weights and biases of neurons $z$ are randomly initialized and are bounded by $\mathbb{E}[||\nabla z(x)||] \leq C$ for some $C > 0$, the average number of linear regions is proportional to the product of the number of neurons and the size of training set, that is,
\begin{equation}\nonumber
	\mathbb{E}[\#\{linear\ regions\ in\ S\}] \approx |S|\cdot T\cdot M
\end{equation}
where $S$ is the training dataset, $T$ is the number of breakpoints in the activation function (for ReLU, $T=1$), and $M$ is the total number of hidden neurons. 

Second, they investigate ReLU neural networks whose input dimensionality exceeds 1, denoted by $\mathcal{N}: \mathbb{R}^d \rightarrow \mathbb{R}\ (d>1)$, and use the volume of boundaries between linear regions in the input space as an estimation of model complexity. That is, 
\begin{equation}\nonumber
	EMC(\mathcal{N}) = \frac{volume_{d-1}(B_{\mathcal{N}} \cap K)}{volume_d(K)}
\end{equation}
where $B_\mathcal{N} = \{x | \nabla \mathcal{N}(x)\text{ is not continuous at } x\}$ represents the boundaries of the linear regions formed by $\mathcal{N}$, $K \in \mathbb{R}^{d}$ is the data distribution. 
They prove that, under the same parameter initialization assumption as $d=1$, the expected value of the volume of linear region boundaries is approximately equal to
\begin{equation}\nonumber
	\mathbb{E}[EMC(\mathcal{N}) ] \approx T\cdot M
\end{equation}
This demonstrates that the average size of the region boundaries depends only on the number of neurons, not on the depth of the model. They conclude that the effective complexity of deep neural networks may be much lower than the theoretical bound. That is, the function learned by deep neural networks may not be more complex than that learned by shallow ones.

\subsection{Discussion}
\label{sec:effectivediscussion}


Effective model complexity is a relatively new, promising and useful problem in deep learning. 
Detecting effective model complexity during training helps to investigate the usefulness of optimization algorithms~\cite{kalimeris2019sgd}, the role of regularizations~\cite{hu2020lann,raghu2017expressive}, and generalization capability~\cite{liang2017fisher,novak2018sensitivity}. 
Furthermore, effective model complexity can be used to describe model compression ratio, since effective model complexity can be considered as a reflection of the information volume in the model~\cite{du2016weight}.
Effective complexity can also be used for model selection and design to balance resource utilization and model performance.

In addition to the effective complexity measures and the high-capacity low-reality phenomenon, there are a series of interesting problems about effective complexity of deep learning models. For example, the cross-model comparison of effective complexity is worth exploring. That is, how to compare the effective complexity of multiple models with different architectures, and how the effective complexity is affected by the choice of different model architectures. 
Moreover, can one specify the granularity of effective complexity measures? Different cases may have different requirements for effective complexity. Correspondingly, the application scopes and granularities of effective complexity measures should be specified and clarified. Typically, effective complexity measure by the number of non-zero parameters is obviously not sufficient to study the optimization process.

\nop{
\medskip

In this section, we review several works that explore the measure of effective complexity~\cite{raghu2017expressive,novak2018sensitivity,hu2020lann,nakkiran2019deep,liang2017fisher} and the high-capacity low-reality fact~\cite{ba2014deep,hanin2019complexity} in the effective complexity of deep learning models. We also discuss the validity of effective complexity, and several unsolved problems. 
}

\section{Application Examples of Deep Learning Model Complexity}
\label{sec:utility}

Model complexity of deep learning has many applications. In this section, we review three interesting applications of deep learning model complexity, namely understanding model generalization capability, model optimization, and model selection and design.

\subsection{Model Complexity in Understanding Model Generalization Capability}

Deep learning models are always over-parameterized, that is, they have far more model parameters than the optimal solutions and the number of training samples. However, it is often found that large over-parameterized neural networks exhibit good generalization capability~\cite{kawaguchi2017generalization,neyshabur2017exploring,zhang2016understanding}. Some studies even find that larger and more complex networks usually generalize better~\cite{novak2018sensitivity}. This observation is in contradiction with the classical notion of function complexity~\cite{novak2018sensitivity} and the well-known Occam's razor~\cite{rasmussen2001occam}, which prefer simple models. What leads to the good generalization capability of over-parameterized deep learning models?

In statistical learning theory, expressive capacity (i.e., hypothesis space complexity) is used to bound generalization error~\cite{mohri2018foundations}. Specifically, let $\mathcal{F}$ be the set of functions representable by a certain model structure. Let $f_{A(D)}$ be a function $f\in \mathcal{F}$ learned by algorithm $A$ on training dataset $D$. Let $E_D(f_{A(D)})$ be the emperical error of $f_{A(D)}$ and $E(f_{A(D)})$ the generalization error of $f_{A(D)}$. The gap between generalization error and emperical error is bounded by
\begin{equation} \label{eq:generalizationerror}
E(f_{A(D)}) - E_D(f_{A(D)}) \leq \sup_{f \in \mathcal{F}} \{E(f) - E_D(f)\}
\end{equation}
The right-hand side can be quantified by analyzing the expressive capacity~(e.g., Rademacher complexity)~\cite{kawaguchi2017generalization}.
For example, Zheng~\textit{et al.}~\cite{zheng2019capacity} analyze generalization error of deep ReLU neural networks using the basis-path norm, a norm measure based on the basis paths. 
Zheng~\textit{et al.}~\cite{zheng2019capacity} suggest that there exist a small group of basis paths in a ReLU neural network, which are linearly independent. Each input-output path of a ReLU neural network can be expressed in the form of multiplication and division of the basis paths. 
Therefore, the basis paths can be used to explain the generalization behavior of ReLU neural networks. 
Zheng~\textit{et al.}~\cite{zheng2019capacity} prove that the generalization error of ReLU neural networks (i.e. the right-hand side of Eq.~(\ref{eq:generalizationerror})) can be bounded by a function of basis-path norm. 

A series of studies investigate model complexity measures that can explain generalization capability of deep learning models~\cite{allen2019learning,liang2017fisher,neyshabur2017exploring,novak2018sensitivity}. 
Neyshabur~\textit{et al.}~\cite{neyshabur2017exploring} suggests that, from the perspective of generalization, a complexity measure should satisfy the following property: a learned model with lower complexity generalizes better.
In particular, they list several requirements which are summarized from observed empirical phenomena and are expected to be satisfied by complexity measures.
\begin{itemize}
	\item With zero training error, a network trained on real labels, which leads to good generalization, is expected to have much lower complexity than a network trained on random labels. 
	
	\item Increasing the number of hidden units or the number of parameters, which leads to a decreased generalization error, is expected to decrease the complexity measure.
	
	\item When training the same architecture on the same training dataset using two different optimization algorithms, if both lead to zero training errors, the model with better generalization is expected to have lower complexity.
\end{itemize}

Based on these desiderata, Neyshabur~\textit{et al.}~\cite{neyshabur2017exploring} investigate several complexity measures including norms~\cite{neyshabur2015norm}, robustness~\cite{xu2012robustness}, and sharpness~\cite{keskar2016large}.  They show that, these measures can meet some of the above requirements, but not all.

Novak~\textit{et al.}~\cite{novak2018sensitivity} define two complexity measures from the perspective of model sensitivity, and identify an empirical correlation between the complexity measures and model generalization capability. They show that operations that lead to poor generalization, such as full batch training, correspond to high sensitivity, and in turn imply high effective model complexity. Similarly, operations that lead to good generalization, such as data augmentation, correspond to low sensitivity, and thus imply low effective model complexity.

Liang~\textit{et al.}~\cite{liang2017fisher} define a complexity measure using the Fisher-Rao norm to investigate model generalization capability. 
They suggest that a complexity measure used to study generalization should satisfy the invariance property. The invariance property requires that the generalization capacity depends on the equivalence classes obtained by deep models. In other words, many different parameterizations may lead to the same prediction. Thus, the specific parameterization of deep models should not affect the generalization and the complexity measure.
They show that the Fisher-Rao norm honors this invariance property and thus is able to explain the generalization capability of deep learning models.

\subsection{Model Complexity in Optimization}

Model optimization is concerned about how and why a neural network model can be successfully trained~\cite{poggio2017theory,sun2019optimization}. Specifically, the optimization of a deep model is to determine model parameters to minimize a loss function in general non-convex. The loss function is typically designed based on the understanding of a problem and the requirements of the model, and thus generally includes a performance measure, which is evaluated on the training set, and other constraint terms~\cite{goodfellow2016deep}.


Model complexity is widely used to provide a metric to make optimization traceable. 
For example, a measure metric of the effective model complexity of neural networks helps to monitor the changes of a model during the optimization process and understand how the optimization process progresses~\cite{hu2020lann,kalimeris2019sgd,nakkiran2019deep,raghu2017expressive}. Such a metric also helps to verify the effectiveness of new improvements of optimization algorithms~\cite{hayou2018selection}. For example, Nakkiran~\textit{et al.}~\cite{nakkiran2019deep} investigate the double descent phenomenon during training using effective complexity measured by the maximal size of the dataset on which zero training error can be achieved. They show that the double descent phenomenon can be represented as a function of the effective complexity. Raghu~\textit{et al.}~\cite{raghu2017expressive} and Hu~\textit{et al.}~\cite{hu2020lann} propose new regularization methods and demonstrate the effectiveness of these regularizations through their impact on complexity. 

The study of model complexity inspires explorations on the effectiveness of optimization approaches. 
Hanin and Rolnick~\cite{hanin2019complexity} use the boundary volumes of linear regions as the complexity measure of ReLU neural networks, and find that during the training, the average boundary volume is always linearly proportional to the number of neurons, irrelevant to the depth of the model. This demonstrates that deep models do not always learn more complex functions than shallow ones, the success of deep learning may be related to optimization algorithms.
Ba and Caruana~\cite{ba2014deep} suggest that the great performance of deep learning may be due to the fact that deep models are easier to train than shallow architectures using the current optimization algorithms~\cite{hanin2019complexity,novak2018sensitivity}. This calls for further exploration of the effectiveness of optimization approaches and the relationship with model structures.

\subsection{Model Complexity in Model Selection and Design} 

Given a specific learning task, how can we determine a feasible model structure for the task? Given a variety of models with different architectures and different complexity, how can we pick the best model from them? This is the model selection and design problem~\cite{murphy2012machine}. 

In general, model selection and design is based on the tradeoff between prediction performance and model complexity~\cite{lim2000comparison,myung2000importance}. On one hand, making predictions with high accuracy is the essential goal of learning a model~\cite{mohri2018foundations}. A model is expected to be able to capture the underlying patterns hidden in the training data and achieve predictions of accuracy as high as possible. In order to represent a large amount of knowledge and obtain high accuracy, a model with a high expressive capacity, a large degree of freedom and a large training set is required~\cite{bengio2011expressive}. To this extent, a model with more parameters and higher complexity is favored. 
On the other hand, an overly complex model may be difficult to train and may incur unnecessary resource consumption, such as storage, computation and time cost~\cite{myung2000importance}. Unnecessary resource consumption should be avoided particularly in practical large scale applications~\cite{hoge2018primer}. To this extent, a simpler model with comparable accuracy is preferred than a more complicated one. 

To maintain a good tradeoff between accuracy and complexity, a model selected is expected to be complex enough to fit the given data and achieve high accuracy. At the same time, the model should not be highly over-complicated. Understanding model complexity and developing an effective complexity measure are the premise for good model selection strategies. 
For instance, Michel and Nouy~\cite{michel2020learning} propose a model selection strategy for tree tensor networks (i.e., sum-product neural networks). Their method combines the empirical risk minimization and model complexity penalty to select a model from a family of models.

Neural architecture search (NAS for short) is a popular solution to deep learning model selection~\cite{laredo2020automatic,liu2018progressive,liu2018darts,zoph2016neural}, which automatically selects a good neural network architecture for a given learning task.
Since an overly complex model may take too-long training time and thus may become a serious obstacle of neural architecture search~\cite{laredo2020automatic,liu2018progressive}, the accuracy-complexity tradeoff is an important consideration in neural architecture search.
Liu~\textit{et al.}~\cite{liu2018progressive} propose Progressive Neural Architecture Search, which searches for convolutional neural network architectures in the increasing order of model complexity. Therefore, Progressive Neural Architecture Search favors low complexity models that meet the requirement on prediction accuracy. 
Laredo~\textit{et al.}~\cite{laredo2020automatic} propose Automatic Model Selection, which searches for fully connected neural networks that yield a good balance between prediction accuracy and model complexity. 

Radosavovic~\textit{et al.}~\cite{radosavovic2019network} investigate the network design spaces of model selection and design approaches. They propose to compare design spaces by contrasting the estimated distributions of model complexity in the network design spaces. Using their proposed method of comparing the distributions of model complexity of network design space,  they investigate several popular NAS approaches, such as ENAS~\cite{pham2018efficient} and DARTS~\cite{liu2018darts}, and find that there are significant differences between the network design spaces of these approaches.

\section{Conclusions and Future Directions}
\label{sec:historyfuture}

In this paper, we survey model complexity in deep learning.
We summarize four aspects affecting deep learning model complexity, and two angles to overview existing studies on deep learning model complexity. 
We discuss the two major problems of deep learning model complexity, namely the model expressive capacity and effective model complexity. 
We overview the state-of-the-art studies on the expressive capacity from four aspects: depth efficiency, width efficiency, expressible functional space, and VC dimension and Rademacher complexity.
We overview the state-of-the-art studies on the effective complexity from two aspects: general measures of effective complexity and the high-capacity low-reality phenomenon.
We discuss the application of deep learning model complexity, especially in generalization capability, optimization, model selection and design.


Model complexity of deep learning is still in its infant stage.  There are many interesting challenges for future works.  

Expressive capacity of deep learning models is a challenging problem.
For example, in most cases, deep learning models are over-parameterized and have sufficient expressive power for given tasks and data. A natural question is what expressive capacity is sufficient for a given task.  In other words, can we obtain a lower bound of expressive capacity of deep learning models that are sufficient for a given task? Does a narrow layer limit the expressive capacity of a model even if the model itself has a large number of parameters? 

Several studies explore the bottleneck of model size (i.e., depth, width) in the expressive capacity. That is, when the model size may become a bottleneck that restricts the expressive capacity. For example, Hanin and Sellke~\cite{hanin2017approximating} and Lu~\textit{et al.}~\cite{lu2017expressive} discover that any deep ReLU network with width constrained by the input dimensionality has very limited expressive power. 
Serra~\textit{et al.}~\cite{serra2018bounding} find that smaller widths in the first few layers of a deep ReLU network cause a bottleneck on the expressive power. 
Kileel~\textit{et al.}~\cite{kileel2019expressive} identify a bottleneck of layer width of deep polynomial networks. In a deep polynomial network, if there is a very narrow layer, no matter how wide the other layers are, the network can never correspond to a convex functional space.  A convex functional space benefits the optimization and makes the model easier to train.
Research on the bottleneck of expressive capacity may help to tackle many other problems, such as model design, model selection, model compression, and pruning. 

Though some progress has been made, effective complexity measures are still a largely under-developed direction in deep learning. 
Comparing to expressive capacity of deep learning models, measuring effective model complexity is even more challenging. Effective complexity measures have to be capable of capturing fine granularity differences between two models, such as the same model architecture with two different optimization algorithms.
Several previous studies~\cite{hu2020lann,liang2017fisher,nakkiran2019deep,novak2018sensitivity,raghu2017expressive} define and explore effective complexity measures mainly from the perspective of piecewise linear property~\cite{hu2020lann,novak2018sensitivity,raghu2017expressive}, Fisher-Rao metric~\cite{liang2017fisher}, or the size of trainable samples~\cite{nakkiran2019deep}. However, the effective complexity measure is still a largely unexplored and valuable direction.

Last, cross-model complexity comparison is a promising direction.
Given several models with different model frameworks and different model sizes, how can we compare their expressive capacity? After these models are trained sufficiently on the same dataset, such as obtaining zero training error, how can we compare their effective complexity and further understand their generalization capability? In these cases, the cross-model complexity comparison is useful. Comparing model complexity crossing different deep models can in general help many problems of deep learning, in particular model selection and design.
However, the exploration of cross-model comparison of expressive capacity or effective complexity of deep learning models is still very limited. 
Khrulkov~\textit{et al.}~\cite{khrulkov2017expressive} compare expressive capacity between shallow FCNNs, CNNs, and RNNs by connecting network architectures to tensor decompositions. 
However, many more sophisticated models are not involved and need to be further explored.




\bibliographystyle{abbrv}
\bibliography{references}

\end{sloppy}
\end{document}